%% file: main.tex
\documentclass[sigconf, review=false]{acmart}

\usepackage{booktabs} 
\usepackage{float}

\setcopyright{cc}
\setcctype{by}

\usepackage{tcolorbox}
\tcbuselibrary{listingsutf8}
\tcbuselibrary{breakable}
\usepackage{multirow}
\usepackage{tabularx}
\usepackage{placeins}

\newcommand{\biryani}{\textit{biryani }}
\newcommand{\Biryani}{\textit{Biryani }}

\acmConference[ICVGIP 2025]{Indian Conference on Computer Vision, Graphics, and Image Processing}{December 17--20, 2025}{Mandi, India}
\acmYear{2025}
\copyrightyear{2025}

\acmBooktitle{Indian Conference on Computer Vision, Graphics, and Image Processing (ICVGIP 2025), December 17--20, 2025, Mandi, India}
\acmPrice{}
\acmDOI{10.1145/3774521.3774596}
\acmISBN{979-8-4007-1930-1/2025/12}

\makeatletter
\renewcommand\@fnsymbol[1]{\ifcase#1\or †\else\@arabic{#1}\fi}
\makeatother

\begin{document}
\title{How Does India Cook \textit{Biryani}?}

\author{C V Rishi}
\authornote{Equal Contribution}
\affiliation{%
  \institution{IIIT Hyderabad}
  \country{India}
}
\email{se23ucse044@mahindrauniversity.edu.in}

\author{Farzana S}
\authornotemark[1]
\affiliation{%
  \institution{IIIT Hyderabad}
  \country{India}
}
\email{farzana.s@research.iiit.ac.in}

\author{Shubham Goel}
\authornotemark[1]
\affiliation{%
  \institution{IIIT Hyderabad}
  \country{India}
}
\email{shubham.goel@students.iiit.ac.in}

\author{Aditya Arun}
\affiliation{%
  \institution{IIIT Hyderabad}
  \country{India}
}
\email{adityaarun1@gmail.com}

\author{C V Jawahar}
\affiliation{%
  \institution{IIIT Hyderabad}
  \country{India}
}
\email{jawahar@iiit.ac.in}

\renewcommand{\shortauthors}{}

\begin{abstract}
\textit{Biryani}, one of India’s most celebrated dishes, exhibits remarkable regional diversity in its preparation, ingredients, and presentation. With the growing availability of online cooking videos, there is unprecedented potential to study such culinary variations using computational tools systematically. However, existing video understanding methods fail to capture the fine-grained, multimodal, and culturally grounded differences in procedural cooking videos. This work presents the first large-scale, curated dataset of \biryani preparation videos, comprising 120 high-quality YouTube recordings across 12 distinct regional styles. We propose a multi-stage framework leveraging recent advances in vision–language models (VLMs) to segment videos into fine-grained procedural units and align them with audio transcripts and canonical recipe text. Building on these aligned representations, we introduce a video comparison pipeline that automatically identifies and explains procedural differences between regional variants. We construct a comprehensive question–answer (QA) benchmark spanning multiple reasoning levels to evaluate procedural understanding in VLMs. Our approach employs multiple VLMs in complementary roles, incorporates human-in-the-loop verification for high-precision tasks, and benchmarks several state-of-the-art models under zero-shot and fine-tuned settings. The resulting dataset, comparison methodology, and QA benchmark provide a new testbed for evaluating VLMs on structured, multimodal reasoning tasks and open new directions for computational analysis of cultural heritage through cooking videos. We release all data, code, and the project website at \url{https://farzanashaju.github.io/how-does-india-cook-biryani/}.
\end{abstract}
%
%

\maketitle

\input{content/introduction}

\input{content/dataset_and_viz}

\input{content/video_seg}
\input{content/video_comparison}

\input{content/qa_gen}
\input{content/applications_and_conclusion}

\bibliographystyle{ACM-Reference-Format}
\bibliography{references}

\appendix
\section{Biryani Categories}
\label{sec:appendix_biryani_categories}

\begin{enumerate}

    \item \textbf{Ambur Biryani:} Originating from Ambur in Tamil Nadu, this version is made with flavorful seeraga samba (Jeeraga Samba) rice, lending a distinct aroma and texture. It is said to have royal roots in the Nawab of Arcot’s kitchens and is typically served with a tangy eggplant curry. 
    
    \item \textbf{Bombay Biryani:} A fusion of Persian, Mughlai, and Maharashtrian styles, Bombay biryani is a flavorful dum-cooked rice dish commonly featuring potatoes and sometimes dried plums, with a milder spice profile. 

    \item \textbf{Dindigul Biryani:} Known particularly as Dindigul Thalapakatti Biryani, it uses seeraga samba rice and bold, tangy flavours—often featuring goat meat. It distinguishes itself through its slow-cooking technique and intense taste profile. 

    \item \textbf{Donne Biryani:} A fragrant South Indian biryani, especially from Bangalore’s “Military Hotel” style, this biryani uses seeraga samba rice and a freshly ground masala paste, served traditionally on a disposable leaf-paper “donne.”  
    \item \textbf{Hyderabadi Biryani:} Hailing from Hyderabad’s Nizam kitchens, this iconic dum-cooked biryani comes in two variants: kachchi (raw marinated meat layered with rice) and pakki (cooked meat). It features basmati rice, meat, spices, saffron, and fried onions.

    \item \textbf{Kashmiri Biryani:} Typically a vegetarian style from Kashmiri Pandit tradition, it’s made without onion or garlic and often includes vegetables, yoghurt, nuts, and fragrant basmati rice—a milder, saffron-infused version. Alternatively, mutton-based Kashmiri biryani includes dry fruits and kewra for a delicate flavour.

    \item \textbf{Kolkata Biryani:} Invented in the 1850s–60s by Nawab Wajid Ali Shah in exile, this biryani incorporated potatoes, eggs, lightly spiced meat, and fragrant rice, adapted from Awadhi-like Mughlai cooking—a lighter, more economical version. 

    \item \textbf{Lucknow Awadhi Biryani:} From Lucknow’s royal kitchens, this biryani is known for its subtle, fragrant flavours, often enhanced with kewra/rose water and saffron. It uses cooked meat layered with al dente rice and steamed “in dum” for refinement. 

    \item \textbf{Malabar Biryani:} A signature of Kerala’s Malabar coast (Kozhikode, Kannur, etc.), this subtly spiced biryani uses short-grain Kaima (Jeerakasala) rice, aromatic ghee, and whole spices like cardamom and cinnamon. It’s mildly sweet, layered with fried onions, cashews, raisins, and cooked on dum for a fragrant, balanced flavour.

    \item \textbf{Mughlai Biryani:} Rooted in Mughal royal cuisine, this biryani is lavish and indulgent—made with basmati rice, meat (or vegetarian), cream, nuts, dried fruits, saffron, and aromatic spices, layered and dum-cooked for rich, creamy indulgence.

    \item \textbf{Sindhi Biryani:} A spicy, tangy, and sweet Pakistani biryani from Sindh, it includes potatoes, tomatoesyoghurtrt, dried plums (aloo bukhara), and a medley of spices. It’s layered and dum-cooked, known for its bold, vibrant flavours.

    \item \textbf{Thalassery Biryani:} A celebrated local variant from Thalassery in North Kerala, this pakki-style biryani separately cooks Kaima rice and meat, then layers them for slow dum cooking. It’s known for its dry, aromatic profile—no oil-heavy richness—and distinctive Kerala spices and ghee-infused rice.  
\end{enumerate}

\section{Prompts Used}
\label{sec:appendix_prompts_used}

\subsection*{Video Segmentation}

\begin{tcolorbox}[colback=blue!5!white, colframe=blue!75!black, title=InternVL-14B Prompt for Segment Analysis, fonttitle=\bfseries]
You are analysing a cooking video.

Please extract information into three clearly labelled bullet-point lists, based strictly on what is visually present in the video frames.

Respond only with the following three sections in this exact order:

\textbf{Ingredients:}
- List all visible ingredients being used (e.g., chopped onions, turmeric powder, rice).

\textbf{Utensils:}
- List all visible cooking tools, vessels, or utensils (e.g., knife, pressure cooker, ladle).

\textbf{Actions:}
- Describe each distinct cooking action as a verb-noun phrase (e.g., chopping onions, frying spices, stirring curry).

\textbf{Important rules:}
- Do NOT include any summary, explanation, or extra commentary.
- Only include items that are visible or implied in the visuals.
- Avoid repeating the same item unless used in a different context.
- Use consistent and specific terms.
\end{tcolorbox}

\begin{tcolorbox}[colback=blue!5!white, colframe=blue!75!black, title=Clustering Decision Prompt, fonttitle=\bfseries]
	You are analysing cooking actions for a biryani recipe classifier. Below is a set of {len(actions)} similar cooking actions that have been grouped:

	{actions\_str}

	Question: Should these actions be split into multiple distinct action classes, or are they similar enough to remain as one group?

	Consider:
	- Are there distinct cooking techniques or steps represented?
	- Would separating them improve classification accuracy for biryani cooking?
	- Are some actions fundamentally different despite semantic similarity?

	Respond with a JSON object containing only:
	{{
			"should\_split": true/false,
		}}
\end{tcolorbox}
\vspace{0.5em}

\vspace{0.5em}
\begin{tcolorbox}[colback=gray!5!white, colframe=gray!75!black, title=Gemini Prompt for Action Verification]
You are an expert in analysing cooking videos. Your task is to determine if a specific action is happening in the provided video frames.

The action to verify is: \textbf{‘[ACTION]’}

 If any \textit{part} of the action is clearly or partially visible—e.g., if the action is “adding turmeric and milk” but only turmeric is visible—answer “yes”.

 Only answer “no” if none of the described actions is visible.

Do not explain. Respond with a single word: “yes” or “no”.
\end{tcolorbox}

\begin{tcolorbox}[colback=blue!5!white, colframe=blue!75!black, title=\textbf{Action Differencing Prompt}, fonttitle=\bfseries]

I am analysing two sets of photos (\texttt{\{total\_frames\}} total) of someone performing the same biryani cooking action: \\

``\texttt{\{action\}}''.

Video A: Photos \texttt{\{clip1\_range\}} \\
Video B: Photos \texttt{\{clip2\_start\}}--\texttt{\{clip2\_end\}} \\[0.5em]

The specific difference to check is: ``\texttt{\{query\_string\}}''. \\
This means I want to determine if Video A shows more of this characteristic compared to Video B. \\[0.5em]

\texttt{\{importance\_context\}} \\[0.5em]

\textbf{Question:} Based on these frames, which video shows more of this difference?
\begin{itemize}
    \item[(a)] Video A
    \item[(b)] Video B
    \item[(c)] They look similar, or it's not clear
    \item[(d)] The videos seem to be irrelevant to the query
\end{itemize}

Be careful: look at the entire set of frames for each video.  
If you are not confident or if the difference is very minor, choose (c). \\[0.5em]

\textbf{Important Guidelines:}
\begin{itemize}
    \item Choose (a) if Video A clearly shows more of the difference than Video B
    \item Choose (b) if Video B clearly shows more of the difference than Video A
    \item Choose (c) if you cannot confidently distinguish between them or they \\ appear similar
    \item Choose (d) if the videos do not relate to the query at all / the action \\ shown is completely different to the cooking action
\end{itemize}

Return JSON:
\begin{verbatim}
{
    "answer": "a|b|c|d",
    "confidence": 1-5,
    "difference_visible": true/false,
    "explanation": "Detailed explanation 
                    of what you observed"
}
\end{verbatim}
\end{tcolorbox}

\subsection*{QA Generation}

The following prompts, templates, and illustrative examples present the full details of the input specifications used in our multi-stage question–answer (QA) generation pipeline. While the main paper outlines the methodology at a conceptual level, this section provides the exact instructions given to language models, along with representative intermediate outputs, to ensure reproducibility and transparency.

To produce segment-level natural language descriptions from 10-second video chunks, InternVL3-14B was guided with instructions emphasising explicit mention of ingredients, utensils, cooking actions, and other visually salient details, while avoiding speculative or unverifiable information.

\begin{tcolorbox}[colback=purple!10,colframe=purple!80,title=Video Captioning Prompt]
\small
Generate a detailed and accurate description of a cooking video segment.\\
Use the following guidelines to craft a clear and complete narrative:

\begin{enumerate}
    \item Describe key visual elements such as ingredients, utensils, appliances, and the appearance of food at different stages of preparation.
    \item Focus on the sequence of actions performed by the cook, including preparation steps (e.g., chopping, mixing, frying), cooking techniques, and transformations in the food (e.g., colour changes, texture changes, boiling).
    \item Highlight interactions between the cook and the ingredients, as well as gestures or tools used.
    \item Emphasise the order of events, transitions between cooking stages, and any significant visual or temporal cues that indicate progress in the recipe.
    \item Ensure the description is thorough yet clear, capturing the essential visual and procedural aspects of the segment to help the viewer understand what is being cooked and how.
\end{enumerate}
\end{tcolorbox}

\begin{figure}[h]
    \centering
    \includegraphics[width=\columnwidth]{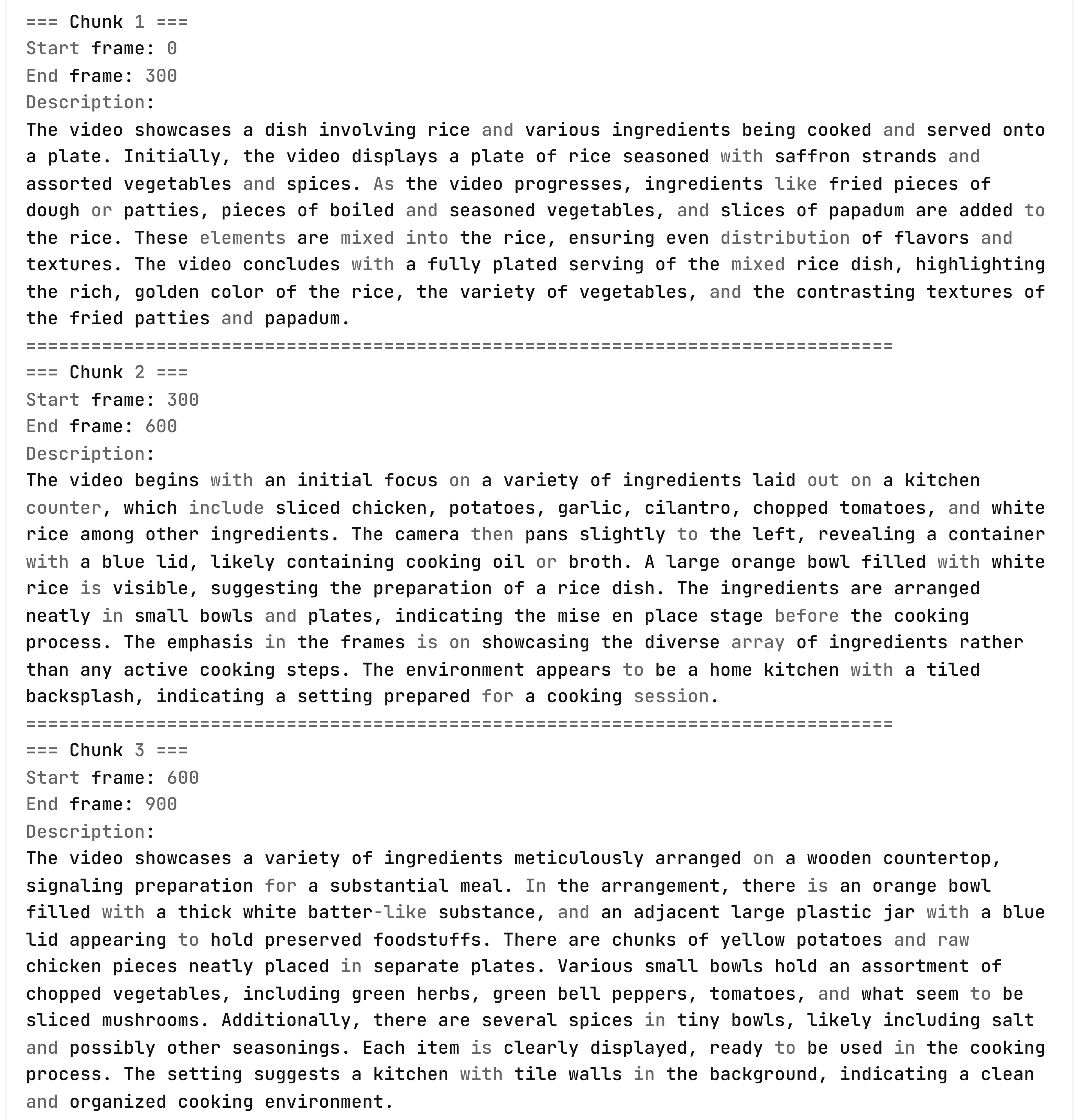}
    \caption{Video Description Example}
    \label{fig:video-description-example}
\end{figure}

Figure \ref{fig:video-description-example} presents an example of a segment-level visual description generated by this captioning stage. The output demonstrates the desired level of detail and specificity, forming the foundation for subsequent summarisation and QA generation.

For the next stage, Gemini-2.0-Flash was prompted to merge all chunk-level descriptions from a video into a coherent, temporally ordered summary. The instructions prioritised preserving event sequence, incorporating visually rich details, and eliminating redundancy, resulting in unified narratives suitable for downstream question generation.

\begin{tcolorbox}[colback=purple!10,colframe=purple!80,title=Video Summarisation Prompt,breakable,]
\small
We split a cooking video into segments and extracted detailed descriptions for each segment. The descriptions for all segments are listed below, in the order they appear in the video. For example, ‘CHUNK: 1’ corresponds to the first video segment.\\

Generate a detailed, step-by-step, and visually rich description of the entire cooking video as a single coherent paragraph, based on all the provided captions. Make sure not to lose any important information. \\
\texttt{"""\\
<segment descriptions>\\
"""}

Use the following instructions to create a clear, complete, and engaging cooking narrative: \\

\begin{enumerate}
    \item Focus on describing key visual details such as the appearance and colours of ingredients, textures, cooking methods, utensils used, hand movements, and how ingredients are combined or transformed during the process.
    \item Preserve the sequence of cooking actions — describe the preparation steps in the order they happen, ensuring the flow matches the progression shown in the captions.
    \item Highlight important details like quantities shown, specific types of ingredients (e.g., green chilli, rice, ginger garlic paste, potatoes), notable textures (e.g., moist, oily, tender), and garnishing or plating details.
    \item Use your reasoning to combine and organise information from all captions into one clear, thorough description. Remove unnecessary repetition and ignore any conflicting or irrelevant details.
    \item Do not mention that the information comes from captions. Present it as a natural, direct description of the video.
    \item Keep it visually descriptive yet easy to understand, almost like explaining the video to someone who can’t watch it.
    \item Finally, use your common sense to conclude what dish is being prepared and summarise how the video showcases its preparation. If the video ends with plating or serving, describe that presentation too.
\end{enumerate}
\end{tcolorbox}

\begin{figure}[h]
    \centering
    \includegraphics[width=\columnwidth]{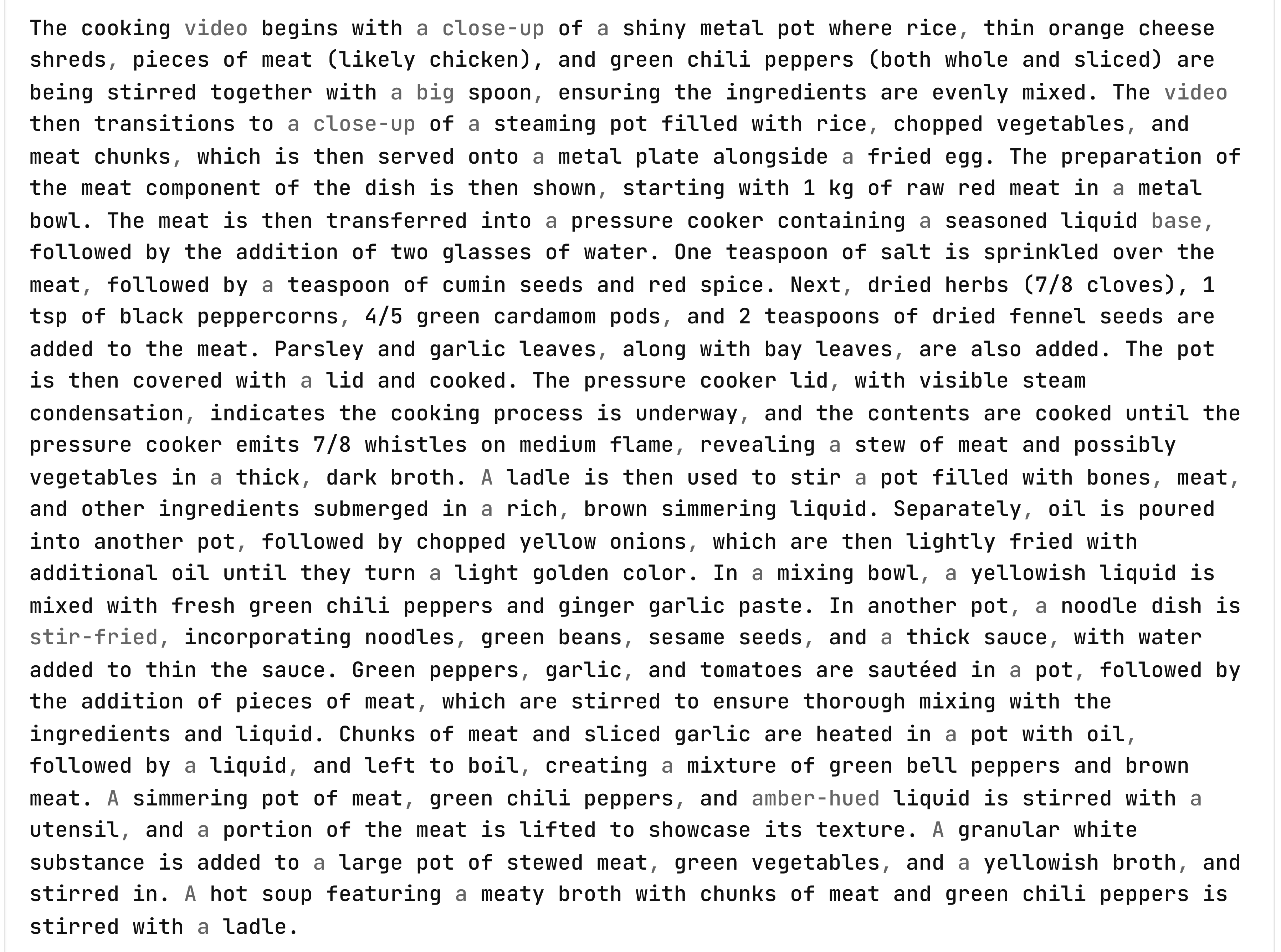}
    \caption{Video Summary Example}
    \label{fig:video-summary-example}
\end{figure}

Figure \ref{fig:video-summary-example} shows an example of a synthesised cooking-video summary produced from multiple segment descriptions. This illustrates how fragmented local observations are transformed into a continuous, recipe-level account.

The pipeline then included an information extraction step in which LLaMA-3-8B-Instruct identified three fixed categories — ingredients, utensils, and cooking actions — from a single segment description.

\begin{tcolorbox}[colback=purple!10,colframe=purple!80,title=Easy QA Generation Prompt]
\small
Video segment description:\\

\texttt{"""\\
<segment description>\\
"""} \\

Answer the following clearly:

\begin{enumerate}
    \item What are the ingredients shown in this segment?
    \item What are the utensils shown in this segment?
    \item What are the cooking actions performed in this segment?
\end{enumerate}
\end{tcolorbox}

Medium-difficulty QA relied on a curated set of question templates covering ingredient usage, step ordering, cooking durations, presentation details, and utensil usage, ensuring questions were grounded in observable visual evidence. These templates were combined with video summaries and transcripts, enabling Gemini-2.0-Flash to generate richer question–answer pairs that integrated multiple sources while avoiding irrelevant or speculative details.

\begin{tcolorbox}[colback=orange!10, colframe=orange!90, title=Medium QA Templates]
\small
\begin{enumerate}
  \item \textbf{What are the primary ingredients used in this recipe?} \\
  \emph{e.g., chicken, rice, yoghurt, spices, onions, tomatoes}

  \item \textbf{In what order are the ingredients added during cooking?} \\
  \emph{e.g., oil → spices → onions → meat → tomatoes → yogurt}

  \item \textbf{Which spices or seasonings are used in this dish?} \\
  \emph{e.g., cumin seeds, coriander powder, garam masala, turmeric, salt}

  \item \textbf{What kind of meat is used in the recipe?} \\
  \emph{e.g., goat, chicken, fish, lamb, beef, none}

  \item \textbf{What is the first step shown in the video?} \\
  \emph{e.g., rinsing and soaking the rice, marinating the meat}

  \item \textbf{What is the last step before serving?} \\
  \emph{e.g., garnishing with fresh coriander and fried onions}

  \item \textbf{How is the meat prepared before cooking?} \\
  \emph{e.g., marinated with yoghurt, turmeric, and chilli powder, layered with meat}

  \item \textbf{What type of pan or vessel is used to cook this dish?} \\
  \emph{e.g., a wide heavy-bottomed metal pot, clay pot, pressure cooker}

  \item \textbf{How long is the rice cooked for?} \\
  \emph{e.g., approximately 15 minutes until tender}

  \item \textbf{Approximately how long does it take to prepare this entire dish?} \\
  \emph{e.g., around 45 minutes total}

  \item \textbf{What does the final dish look like?} \\
  \emph{e.g., orange-red rice with chicken pieces and green garnish}

  \item \textbf{What is used to garnish the dish before serving?} \\
  \emph{e.g., chopped coriander leaves, fried onions, lemon slices}

  \item \textbf{Does the dish appear to be spicy?} \\
  \emph{e.g., yes, it looks spicy due to the visible rechillili oil}

  \item \textbf{When is the rice mixed with the meat or gravy?} \\
  \emph{e.g., after the meat is cooked for 15 minutes}

  \item \textbf{Is the dish served with any accompaniments?} \\
  \emph{e.g., onion raita, boiled eggs, salad}
\end{enumerate}
\end{tcolorbox}

Below is the full prompt provided to Gemini-2.0-Flash for medium-level QA generation. The instructions integrate video summaries with audio transcripts, combine template-guided and model-generated questions, and require answers grounded in the complete cooking process.

\begin{tcolorbox}[colback=purple!10, colframe=purple!80, title=Medium QA Generation Prompt, breakable]
\small
You are an expert in analysing cooking videos, with extensive knowledge of culinary techniques, ingredients, and food presentation across various regional cuisines in India.

You are provided with a detailed textual description of the cooking video and the full transcript of the spoken narration. This data includes step-by-step cooking processes, mentions of ingredients, utensils, cooking durations, and visual cues — but you do not have access to the actual video. \\

\textbf{Task:} \\ \\
- Identify and describe the key cooking processes, ingredients, and presentation details discussed in the textual description and summary. (The key cooking process refers to the main focus of the video that is highlighted in the provided text.)

- Generate relevant Question-Answer (QA) pairs by carefully analysing the textual description and summary of the cooking video.

- In addition to using the provided template questions, feel free to create additional QA pairs that are contextually appropriate based on the content.

Below is a set of template questions for forming QA pairs:
(Adapt these or create new ones depending on the content.) \\

\texttt{"""\\
<templates>\\
"""} \\

\textbf{Instructions:} \\ \\
- DO NOT mention the video summary or transcript directly when answering the questions. Avoid phrases like: “based on the description,” “according to the text,” “as mentioned,” or references to captions that imply the answer was derived from the provided text. Instead, present the information as if it is directly inferred from watching the video.

- Do not explain or justify how the answer was obtained.

- You may choose to omit details that seem irrelevant to the cooking process or final dish.

- Keep all answers concise, and highlight important keywords using bold formatting.

- If a particular question does not apply to the video, simply do not generate a QA pair for it.

- Focus on content directly relevant to the cooking process, ingredients, or presentation. Ignore unrelated background commentary. \\

\textbf{Output Format:}
\begin{verbatim}
{
  "Summary": "",
  "QA_pairs": [
    {"Q": "", "A": ""},
    {"Q": "", "A": ""},
    {"Q": "", "A": ""},
    {"Q": "", "A": ""}
  ]
}
\end{verbatim}

\textbf{Video description:} \\

\texttt{"""\\
<video description>\\
"""}

\textbf{Transcript:} \\

\texttt{"""\\
<transcript>\\
"""}

\end{tcolorbox}

The next stage involved creating multimodal summaries by combining detailed visual descriptions with transcribed spoken instructions. These summaries captured both appearance and process details, incorporating cooking tips, quantities, and sequencing from the narration.

\begin{tcolorbox}[colback=purple!10,colframe=purple!80,title=Multimodal Summarisation Prompt, breakable]
\small
We have split a cooking video into visual segments and extracted detailed descriptions from the video frames for each segment. Separately, we also generated a full transcript of the audio narration spoken in the video.
\\

Your task is to produce a comprehensive, visually and verbally rich summary of the entire cooking video by carefully combining information from both the visual descriptions and the audio transcript.
\\

\textbf{Video description from visual frames:} \\

\texttt{"""\\
<video description>\\
"""}

\textbf{Transcript of the audio narration:} \\

\texttt{"""\\
<transcript>\\
"""}

Use the following instructions to create a clear, complete, and engaging cooking video summary:
 \\

\begin{enumerate}
    \item Use the video summaries from frames to describe key visual details such as the appearance and colours of ingredients, textures, cooking methods, utensils, hand movements, how ingredients are layered or transformed, and plating or serving scenes.
    \item Use the transcript of the audio narration to incorporate spoken explanations, cooking tips, quantities, and verbal emphasis on techniques or ingredient choices.
    \item Ensure the cooking steps are described in the correct sequence, matching the flow shown across the video segments and the spoken instructions.
    \item Highlight important specifics like ingredient types (e.g., green chillies, basmati rice, ginger garlic paste, bone-in chicken), notable textures (e.g., golden fried onions, oily masala, tender meat), quantities or approximate amounts mentioned, and final garnishing or plating details.
    \item Merge and organise all this information into one clear, thorough, and engaging description, removing unnecessary repetition and ignoring conflicting or irrelevant details.
    \item Do not mention captions, transcripts, or segments explicitly. Present it as if you are naturally describing what is happening in the video.
    \item Keep the narrative vivid and easy to understand, as if explaining the video to someone who cannot watch it.
    \item Conclude by summarising what dish is being prepared and how the video showcases its preparation, including the final presentation if shown.
\end{enumerate}
\end{tcolorbox}

\begin{figure}[h]
    \centering
    \includegraphics[width=\columnwidth]{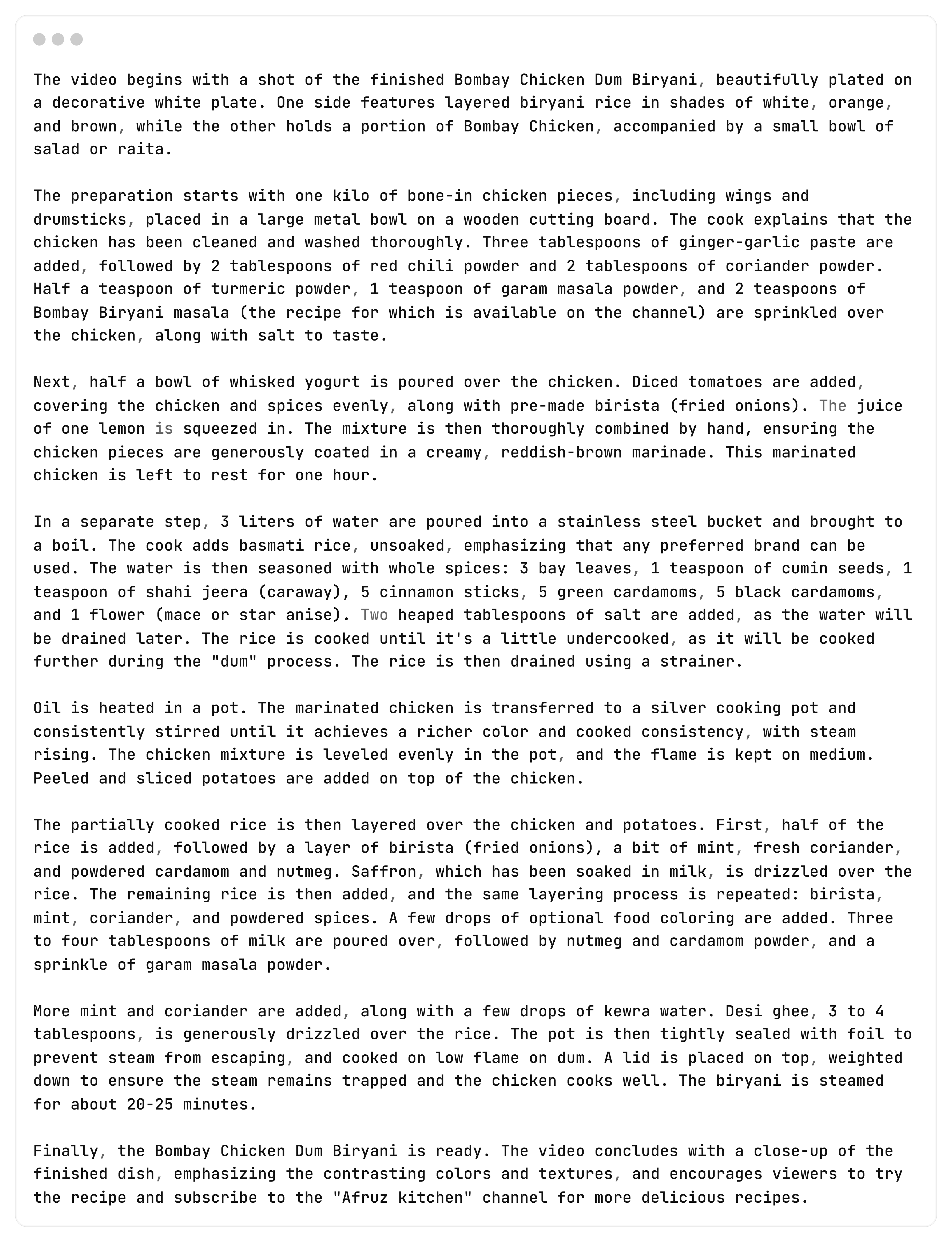}
    \caption{Multimodal Summary Example}
    \label{fig:multimodal-summary-example}
\end{figure}

Figure \ref{fig:multimodal-summary-example} provides an example of such a multimodal summary, illustrating how complementary visual and auditory information is integrated into a single, highly detailed representation of the cooking process.

Finally, reasoning-intensive QA generation was carried out by comparing and contrasting multiple multimodal summaries. A dedicated set of high-level question templates supported cross-video analysis, addressing similarities and differences in ingredients, techniques, spice usage, preparation order, and presentation styles. This stage required synthesis across multiple examples to produce challenging, reasoning-oriented question–answer pairs.

\begin{tcolorbox}[colback=orange!10, colframe=orange!90, title=Hard QA Templates]
\small
\begin{enumerate}

  \item \textbf{Which ingredient is common across all the recipes shown?} \\
  \emph{e.g., onions are used in all three dishes}

  \item \textbf{Which dish uses the highest variety of spices?} \\
  \emph{e.g., the Hyderabad biryani uses 7 different spices, more than the others}

  \item \textbf{Which recipe takes the longest time to prepare?} \\
  \emph{e.g., the Lucknow biryani takes approximately 1 hour}

  \item \textbf{Which of the recipes do not include yoghurt as an ingredient?} \\
  \emph{e.g., only the Ambur biryani skips yoghurt}

  \item \textbf{In which video is rice boiled separately before adding to the meat, unlike in the others?} \\
  \emph{e.g., the Lucknow recipe}

  \item \textbf{Which recipe appears thspiciestcy?} \\
  \emph{e.g., the Andhra biryani looks deep red from heavy chilli usage}

  \item \textbf{In which video does the cook add the meat later in the cooking process compared to the others?} \\
  \emph{e.g., the Kerala biryani adds meat after vegetables}

  \item \textbf{Which videos are the most different from each other?} \\
  \emph{e.g., the Kerala and Hyderabad biryanis differ greatly in cooking method and garnish}

  \item \textbf{Which videos are the most similar to each other?} \\
  \emph{e.g., the Ambur and Tamil Nadu biryanis are nearly identical}
  
\end{enumerate}
\end{tcolorbox}

Below is the final prompt used with Gemini-2.5-Flash to generate reasoning-intensive QA pairs requiring the integration of information from multiple multimodal video summaries.
It instructs the model to identify and synthesise cross-video patterns and distinctions that cannot be inferred from a single source.

\begin{tcolorbox}[colback=purple!10, colframe=purple!80, title=Hard QA Generation Prompt, breakable]
\small
You are an expert in analysing cooking videos, with extensive knowledge of culinary techniques, ingredients, and food presentation across various regional cuisines in India.

You are provided with textual summaries of multiple cooking videos. These summaries include step-by-step actions, mentions of ingredients, utensils, and visual cues — but you do not have access to the actual videos themselves. \\

\textbf{Task:} \\ \\
- Carefully compare, contrast, and synthesise the details across these multiple videos to identify key differences, similarities, and unique aspects. This includes analysing cooking processes, ingredients, preparation times, spice usage, visual appearance, and sequencing of steps.

- Generate high-level, challenging Question-Answer (QA) pairs that require reasoning across these multiple videos, not just describing a single video.

- Use the provided set of question templates to guide your QA generation. You may also create additional multi-video QA pairs if they are insightful. \\

Below is a set of template questions for forming QA pairs:
(Adapt these or create new ones depending on the content.) \\

\texttt{"""\\
<templates>\\
"""} \\

\textbf{Instructions:} \\ \\
- Do not mention the video summaries or textual descriptions directly when answering the questions. Avoid phrases like: “based on the description,” “according to the text,” “as mentioned,” or references to captions that imply the answer was derived from the provided summaries. Instead, present the information as if it is directly inferred from watching the videos.

- Do not explain or justify how the answer was obtained.

- Keep all answers concise, and highlight important keywords using bold formatting.

- If a particular question does not apply to this set of videos, simply do not generate a QA pair for it.

- Focus on content directly relevant to the cooking processes, ingredients, or comparative aspects. Ignore unrelated background commentary. \\

\textbf{Output Format:}
\begin{verbatim}
{
  "Summary": "",
  "QA_pairs": [
    {"Q": "", "A": ""},
    {"Q": "", "A": ""},
    {"Q": "", "A": ""},
    {"Q": "", "A": ""}
  ]
}
\end{verbatim}

\textbf{Video summaries:} \\

\texttt{"""\\
<video summaries>\\
"""}

\end{tcolorbox}

\section{Question Answer Examples}
\label{sec:appendix_qa_examples}

This section presents representative question–answer (QA) pairs from the easy, medium, and hard difficulty tiers of the dataset. These examples illustrate how the prompts, templates, and generation procedures described in Section S2 are applied in practice, highlighting the distinct characteristics and reasoning demands of each difficulty level.

The easy tier focuses on localised, segment-level visual observations. Questions are designed to be direct and unambiguous, answerable from a short video segment without requiring broader temporal or cross-modal reasoning.

\begin{figure}[h]
\centering
\includegraphics[width=\columnwidth]{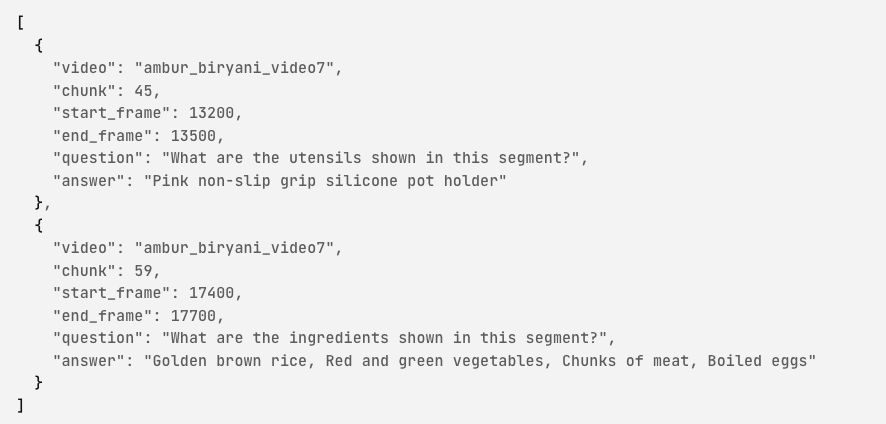}
\caption{Easy Example 1}
\label{fig:easy-1}
\Description{Example JSON annotations from the dataset. Each entry corresponds to a 10-second video segment and contains metadata including the video identifier, chunk index, start and end frame numbers, a natural language question about the segment, and its corresponding answer. In the first example, the system identifies the utensil “Pink non-slip grip silicone pot holder.” In the second example, it lists detected ingredients: “Golden brown rice, Red and green vegetables, Chunks of meat, Boiled eggs.”}

\end{figure}

\begin{figure}[h]
\centering
\includegraphics[width=\columnwidth]{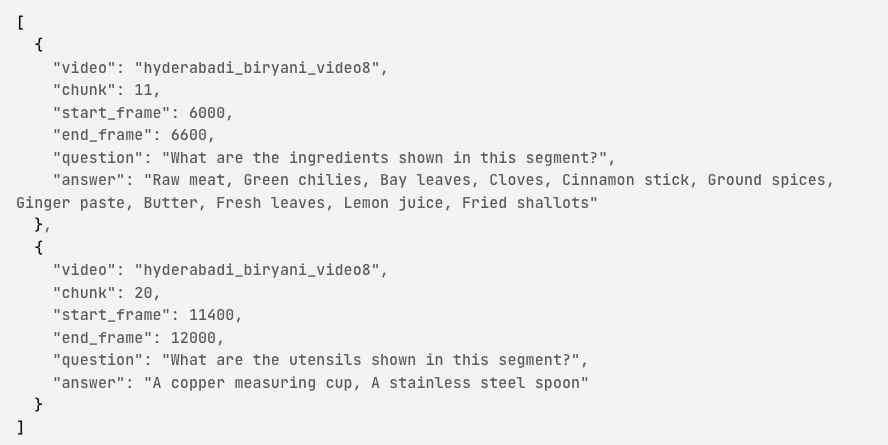}
\caption{Easy Example 2}
\Description{Example JSON annotations from the dataset. Each object contains the video identifier, chunk index, start and end frame numbers, a question about the segment, and its corresponding answer. In the first case, the system lists ingredients such as “Raw meat, Green chilies, Bay leaves, Cloves, Cinnamon stick, Ground spices, Ginger paste, Butter, Fresh leaves, Lemon juice, Fried shallots.” In the second case, it identifies utensils: “A copper measuring cup, A stainless steel spoon.”}

\label{fig:easy-2}
\end{figure}

\begin{figure}[h]
\centering
\includegraphics[width=\columnwidth]{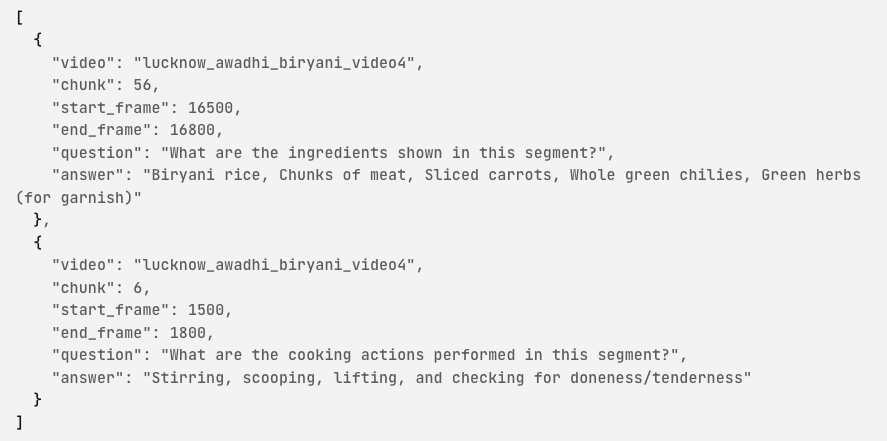}
\caption{Easy Example 3}
\label{fig:easy-3}
\Description{Example JSON annotations from the dataset for Lucknow Awadhi biryani videos. Each entry includes the video identifier, chunk index, start and end frame numbers, a question about the segment, and its answer. The first entry lists ingredients such as “Biryani rice, Chunks of meat, Sliced carrots, Whole green chilies, Green herbs (for garnish).” The second entry describes cooking actions performed in the segment: “Stirring, scooping, lifting, and checking for doneness/tenderness.”}

\end{figure}

Figures~\ref{fig:easy-1}–\ref{fig:easy-3} showcase three easy-tier examples, each containing concise, factual questions about ingredients, utensils, or cooking actions visible within a specific segment.

The medium tier integrates information from entire video summaries and transcripts. These questions require temporal sequencing, recognition of ingredient roles, and interpretation of the overall cooking process.

\begin{figure}[h]
\centering
\includegraphics[width=\columnwidth]{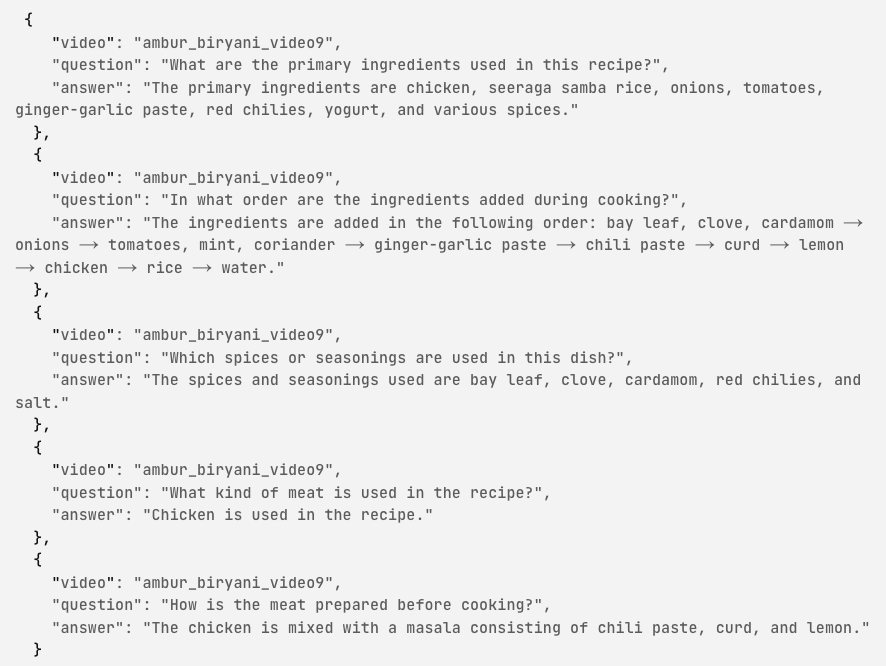}
\caption{Medium Example 1}
\Description{Example JSON annotations for an Ambur biryani recipe. Each entry contains the video identifier, a question about the recipe, and its corresponding answer. The questions address primary ingredients, cooking order, spices used, meat type, and meat preparation method. Answers specify that the dish includes chicken, seeraga samba rice, onions, tomatoes, ginger-garlic paste, red chilies, yogurt, and spices such as bay leaf, clove, and cardamom. The cooking sequence and preparation details, including mixing the meat with chili paste, curd, and lemon, are also documented.}

\label{fig:medium-1}
\end{figure}

\begin{figure}[h]
\centering
\includegraphics[width=\columnwidth]{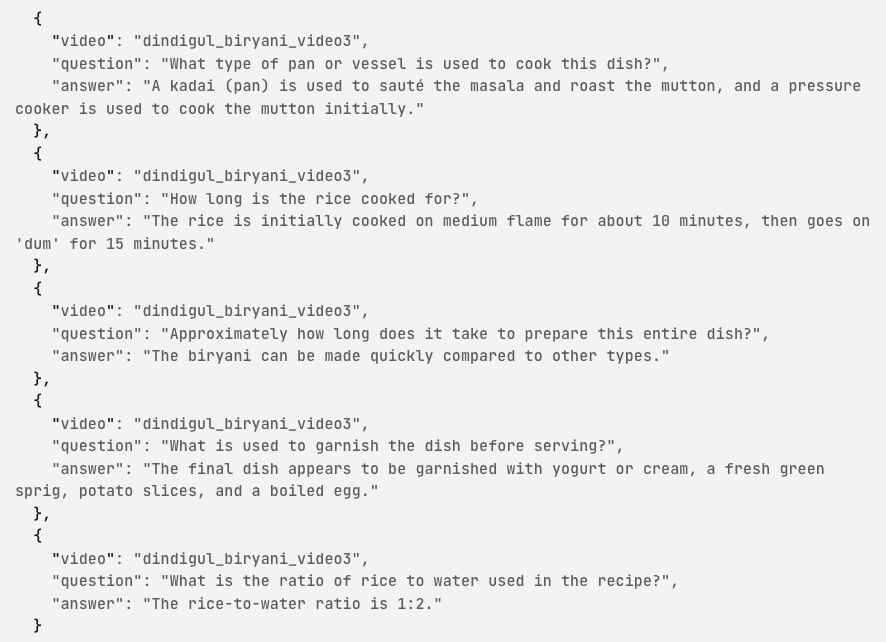}
\caption{Medium Example 2}
\Description{Example JSON annotations for a Dindigul biryani recipe. Each entry records the video identifier, a question about the cooking process, and the corresponding answer. The annotations describe the cooking vessels used, such as a kadai for sautéing and a pressure cooker for cooking mutton, as well as rice cooking times, which include an initial medium flame stage and a 'dum' stage. Additional entries provide information on total preparation time, garnishing ingredients such as yogurt, cream, fresh green sprig, potato slices, and boiled egg, and the specified rice-to-water ratio of 1:2.}

\label{fig:medium-2}
\end{figure}

\begin{figure}[h]
\centering
\includegraphics[width=\columnwidth]{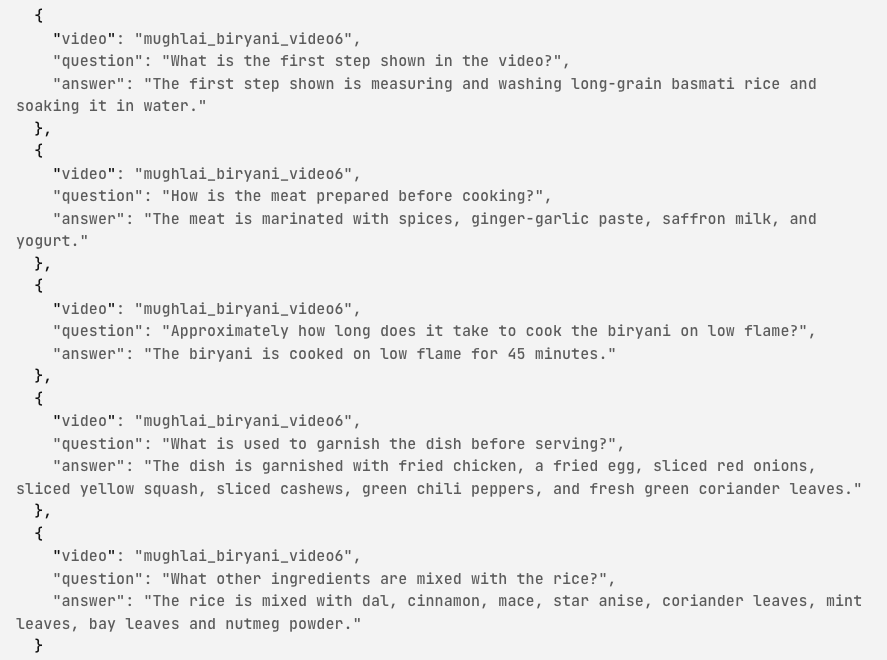}
\caption{Medium Example 3}
\Description{Example JSON annotations for a Mughlai biryani recipe. Each entry lists the video identifier, a cooking-related question, and the corresponding answer. The annotations describe sequential cooking steps, starting with measuring and washing long-grain basmati rice, followed by marinating meat with spices, ginger-garlic paste, saffron milk, and yogurt. Additional entries specify a 45-minute low-flame cooking time, garnishing with fried chicken, a fried egg, sliced vegetables, nuts, green chili peppers, and fresh coriander leaves. The final note details additional ingredients mixed with rice, including dal, cinnamon, mace, star anise, coriander leaves, mint leaves, bay leaves, and nutmeg powder.}

\label{fig:medium-3}
\end{figure}

Figures~\ref{fig:medium-1}–\ref{fig:medium-3} illustrate medium-tier examples, where answering requires synthesising information across multiple steps of preparation while remaining grounded in observable content.

The hard tier requires multi-video comparative and contrastive reasoning. These questions cannot be answered from a single video alone; they demand integration of information across multiple cooking demonstrations to identify similarities, differences, and unique patterns.

\begin{figure}[h]
\centering
\includegraphics[width=\columnwidth]{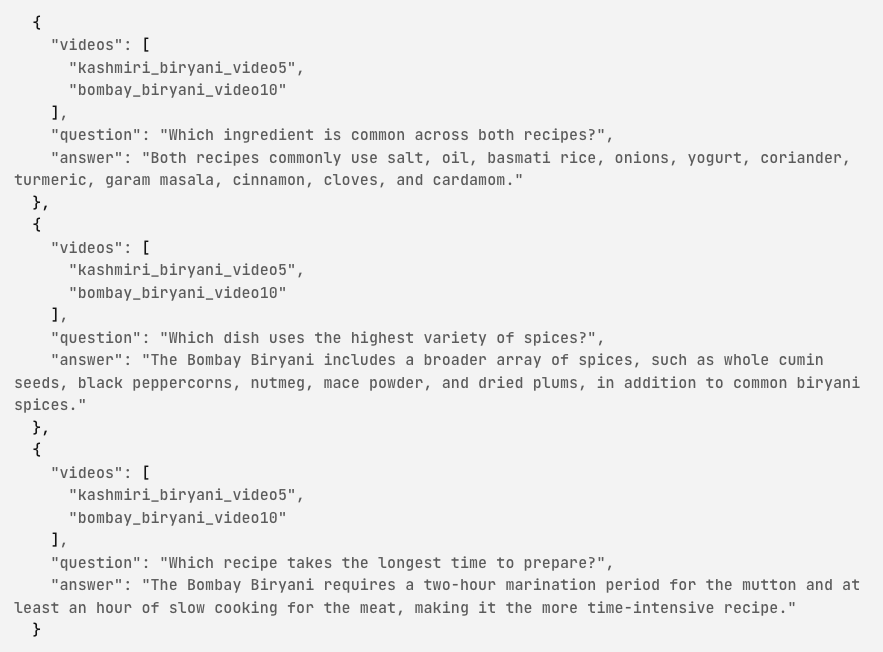}
\caption{Hard Example 1}
\Description{Example JSON annotations comparing Kashmiri and Bombay biryani recipes. Each entry contains a list of video identifiers, a comparative cooking question, and the answer. The comparisons highlight shared ingredients such as salt, oil, basmati rice, onions, yogurt, coriander, turmeric, garam masala, cinnamon, cloves, and cardamom. Differences include the Bombay Biryani’s use of a wider range of spices, including whole cumin seeds, black peppercorns, nutmeg, mace powder, and dried plums. Preparation time is also contrasted, with Bombay Biryani requiring a two-hour marination and at least one hour of slow cooking, making it more time-intensive than the Kashmiri version.}

\label{fig:hard-1}
\end{figure}

\begin{figure}[h]
\centering
\includegraphics[width=\columnwidth]{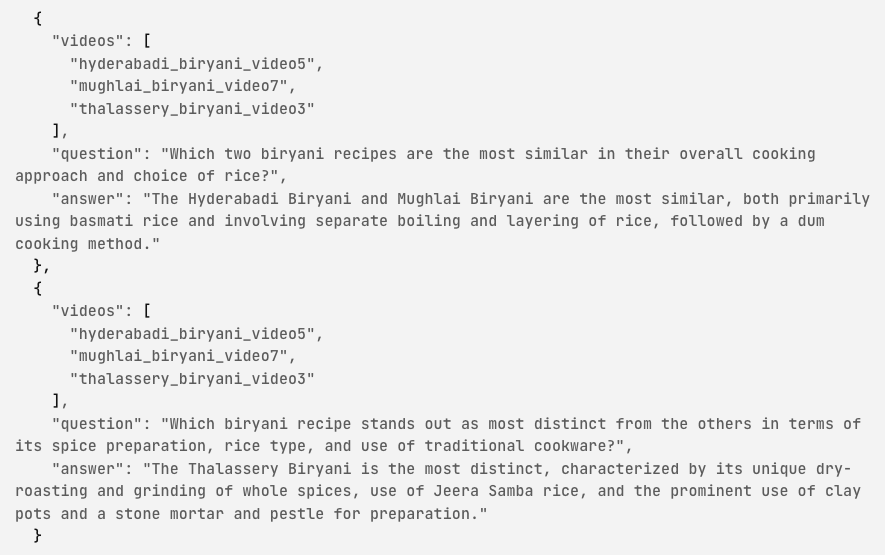}
\caption{Hard Example 2}
\Description{Figure 11: Hard Example 2 shows two JSON objects that compare biryani recipes. 
The first object compares \texttt{Hyderabadi Biryani} and \texttt{Mughlai Biryani}, highlighting their similarity in using basmati rice, separate boiling and layering of rice, followed by a dum cooking method. 
The second object identifies \texttt{Thalassery Biryani} as the most distinct, citing its unique dry-roasting and grinding of whole spices, use of Jeerakasala rice, and the prominent use of clay pots and a stone mortar and pestle for preparation.}

\label{fig:hard-2}
\end{figure}

\begin{figure}[h]
\centering
\includegraphics[width=\columnwidth]{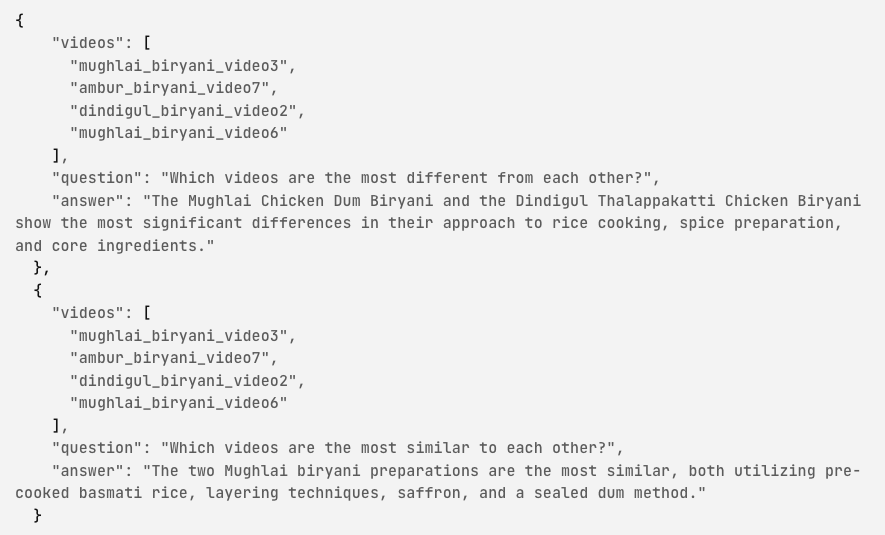}
\caption{Hard Example 3}
\Description{Figure 12: Hard Example 3 contains two JSON objects that compare biryani videos. 
The first object states that \texttt{Mughlai Chicken Dum Biryani} and \texttt{Dindigul Thalappakatti Chicken Biryani} differ the most, showing significant variations in rice cooking, spice preparation, and meat cooking methods. 
The second object finds \texttt{Ambur Biryani} and \texttt{Mughlai Biryani} to be the most similar, as both involve pre-cooked basmati rice, layering techniques, saffron, and sealing the dish during simmering.}

\label{fig:hard-3}
\end{figure}

\begin{figure}[h]
\centering
\includegraphics[width=\columnwidth]{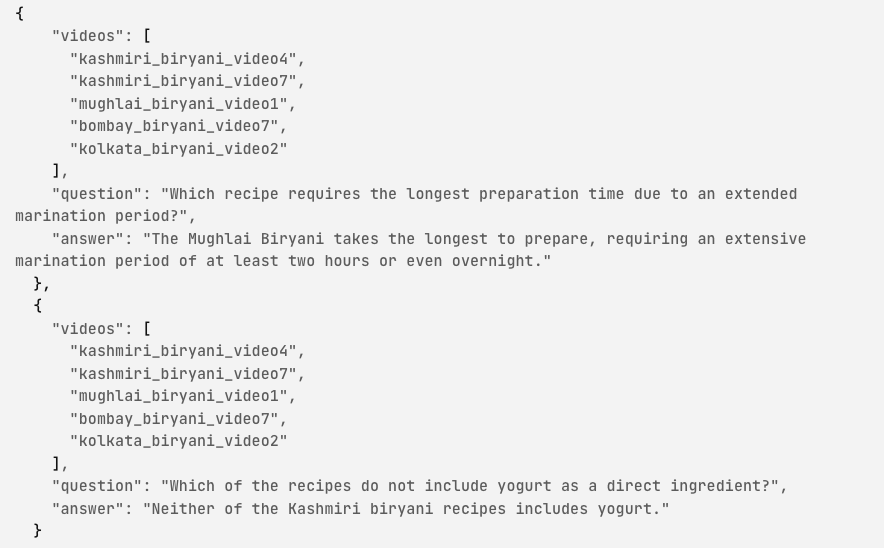}
\caption{Hard Example 4}
\Description{Figure 13: Hard Example 4 presents two JSON objects comparing multiple biryani recipes. 
The first object identifies the \texttt{Mughlai Biryani} as requiring the longest preparation time due to an extensive marination period of at least two hours, or even overnight. 
The second object notes that neither of the \texttt{Kashmiri Biryani} recipes includes yogurt as a direct ingredient.}
\label{fig:hard-4}
\end{figure}

Figures~\ref{fig:hard-1}–\ref{fig:hard-4} present four examples from this tier, demonstrating reasoning over ingredient variations, cooking methods, spice usage, preparation order, and presentation styles across different recipes.

\section{Evaluation Metrics}
\label{sec:appendix_eval_metrics}

\subsection*{BLEU}

The Bilingual Evaluation Understudy (BLEU) metric is an algorithm used to assess the quality of text generated by machine translation from one natural language to another. Its core principle is that the closer a machine’s translation is to that of a skilled human translator, the higher its quality. Developed at IBM in 2001, BLEU was among the first metrics to demonstrate a strong correlation with human quality judgments and remains a widely used, low-cost automatic evaluation method.

BLEU computes scores for individual translated segments—typically sentences—by comparing them against one or more high-quality reference translations. These segment-level scores are then averaged across the entire corpus to estimate overall translation quality. The metric does not account for intelligibility or grammatical accuracy.

The BLEU score ranges from 0 to 1, with higher values indicating greater similarity to the reference translations. A score of 1 is rare even among human translations, as it requires an exact match with a reference. Consequently, a perfect score is not necessary to indicate high quality. 

\subsection*{ROUGE-L}

ROUGE (Recall-Oriented Understudy for Gisting Evaluation) is a widely adopted framework for evaluating the quality of automatically generated summaries—and occasionally translations—by measuring their similarity to one or more human reference texts. The resulting scores range from 0 to 1, with higher values indicating greater alignment with the references.

Among the ROUGE variants, ROUGE-L distinguishes itself by leveraging the Longest Common Subsequence (LCS) between the candidate and reference texts, thereby capturing sentence-level structural similarity rather than merely local n-gram matches. It calculates recall as the ratio of LCS length to the total length of the reference, precision as the ratio of LCS length to the total length of the candidate, and combines these measures via an $F_1$ score.

ROUGE-L’s ability to reward the preservation of word order and coherence makes it particularly useful for assessing the structural fidelity of condensed text. For instance, even when individual words match, a summary with a disrupted sequence will receive a lower ROUGE-L score compared to one that maintains the original flow, highlighting its sensitivity to sentence structure.

\subsection*{BERTScore}

BERTScore is an advanced evaluation metric introduced in 2019 for assessing the quality of machine-generated text by leveraging contextual embeddings derived from pre-trained models like BERT. Unlike traditional evaluation methods such as BLEU or ROUGE, which rely on surface-level word or n-gram matching, BERTScore evaluates semantic similarity through token-level cosine similarity in the embedding space.

The mechanism operates by embedding each token of both the candidate and reference texts using a BERT-based model. It then computes the cosine similarity between all token pairs, using a greedy matching strategy: each candidate token aligns with the most semantically similar reference token for precision, and vice versa for recall. These scores are then harmonised into an F1 measure; optional enhancements such as inverse document frequency (IDF) weighting or baseline rescaling can be applied.

Empirical validation has shown that BERTScore correlates more strongly with human evaluations across various text generation tasks—such as machine translation, summarisation, and image captioning—than traditional metrics. It is particularly effective at capturing semantic equivalence in cases involving paraphrasing or lexical variation.

By focusing on contextual understanding rather than exact token overlap, BERTScore provides a more nuanced and human-aligned evaluation of generated language, making it especially valuable in modern NLP and generative model assessments.

\section{Video Segmentation}
\label{sec:appendix_video_segmentation}

\subsection*{Action clustering}

Direct application of InternVL-14B across thousands of segments yields detailed action descriptions that often vary lexically despite being semantically identical. To address this redundancy, we employed an agglomerative clustering with average linkage on action phrase embeddings generated using the all-MiniLM-L6-v2 SentenceTransformer model. We used a cosine distance of $0.3$ to merge clusters; will no pairs fall below this threshold, we then pick a representative phrase to be the action label.

This clustering process significantly reduces the action vocabulary while preserving semantic diversity. 

The initial action detection stage produced a highly granular label space with 10,481 unique action classes. After applying the action clustering process, this number was reduced to 2,187 canonicalised action classes, representing a 79.1\% reduction while greatly improving consistency in labelling. 

\subsection*{Temporal Merging}

To further enhance temporal coherence, we implemented a clip merging procedure to address fragmentation where identical actions span consecutive temporal segments. This temporal merging process significantly reduced fragmentation in the video segmentation. Across all videos, the number of timestamped clips decreased from 16,761 before merging to 14,479 after merging, representing a 13.6\% reduction in segment count while preserving full action coverage.

\begin{table}[h]
    \centering
    \caption{Action clustering and temporal merging statistics showing significant consolidation in both label space and temporal segmentation}
    \label{tab:clustering_stats}
    \begin{tabular}{lccc}
        \toprule
        \textbf{Process} & \textbf{Before} & \textbf{After} & \textbf{Reduction (\%)} \\
        \midrule
        Action clustering & 10,481 classes & 2,187 classes & 79.1 \\
        Temporal merging  & 16,761 clips   & 14,479 clips  & 13.6 \\
        \bottomrule
    \end{tabular}
\end{table}

\subsection*{Example Data Representation}
To illustrate how our dataset is structured, we provide two representative JSON snippets.  
The first shows a \textbf{10-second temporal segment} annotated with ingredients, utensils, and actions.  
The second shows an \textbf{action-to-timestamp mapping}, where semantically similar action phrases are clustered, and each cluster contains all associated video clips.

\begin{tcolorbox}[colback=gray!5!white,colframe=black,title=10-second Segment Annotation]
\footnotesize
\begin{verbatim}
{
  "timestamp": "59-69",
  "title": "Hyderabadi Chicken Dum Biryani #biryani",
  "url": "https://www.youtube.com/watch?v=BIXMwLFCboA&t=59s",
  "ingredients": [
    "Mint Leaves",
    "Coriander Leaves",
    "Kesar Milk",
    "Kewra & Rose Water",
    "Ghee"
  ],
  "utensils": [
    "Large cooking pot or bowl",
    "Orange cup",
    "Metal cup"
  ],
  "actions": [
    "Adding mint leaves to rice",
    "Adding coriander leaves to rice",
    "Pouring kesar milk over rice",
    "Pouring kewra and rose water over rice",
    "Pouring ghee over rice"
  ]
}
\end{verbatim}
\end{tcolorbox}

\begin{tcolorbox}[colback=gray!5!white,colframe=black,title=Action-to-Timestamped Clips Mapping]
\footnotesize
\begin{verbatim}
"adding bay leaves to the grinder": {
  "phrases": [
    "adding bay leaves to the grinder",
    "placing bay leaf in the spice grinder"
  ],
  "clips": [
    {
      "url": "https://www.youtube.com/watch?v=hgI4wV_WoVs&t=80s",
      "timestamp": "80-90",
      "biryani": "dindigul_biryani",
      "video": "video10"
    },
    {
      "url": "https://www.youtube.com/watch?v=5Zra4nFepRg&t=139s",
      "timestamp": "139-149",
      "biryani": "dindigul_biryani",
      "video": "video1"
    }
  ]
}
\end{verbatim}
\end{tcolorbox}

\noindent
These structured annotations enable fine-grained temporal localisation of cooking actions,  
association with relevant ingredients and utensils, and grouping of semantically similar actions across different videos.  
This organisation supports multimodal reasoning tasks such as step retrieval, ingredient localisation, and cross-video action comparison.

\subsection*{Verification Workflow}
We compile candidate segments grouped by canonical action (e.g., “marinating chicken,” “adding whole spices”), each stored with metadata for action label, video URL, local file path, timestamps (in seconds), \biryani type, and video index. For each 10–30 s segment, we sample up to 20 evenly spaced RGB frames using OpenCV to ensure temporal coverage while controlling input size. These frames are paired with a structured natural language prompt asking Gemini to confirm whether the specified action occurs, where partial or incomplete visibility counts as valid evidence. We query Gemini 2.5 Flash Lite with low temperature for deterministic yes/no outputs, then parse responses as \textit{Correct} for “Yes,” \textit{Incorrect} for “No,” and \textit{Error} for ambiguous or API failures.

\subsection*{Implementation Details}

The complete video segmentation pipeline was executed on NVIDIA A40 GPUs with 48GB VRAM, requiring approximately 12 hours of computation time. InternVL-14B~\cite{chen2023internvl} processed 14,470 video segments across all \textit{biryani} varieties, while the clustering phase operated on the resulting action embeddings using scikit-learn's agglomerative clustering implementation~\cite{Pedregosa2011scikit-learn}.


\begin{table*}[h]
    \centering
    \caption{Implementation details for video segmentation pipeline components showing computational requirements and processing scope}
    \label{tab:segmentation_implementation}
    \begin{tabular}{llll}
        \toprule
        \textbf{Component} & \textbf{Model} & \textbf{Processing Scope} & \textbf{Compute Requirements} \\
        \midrule
        Action detection   & InternVL-14B           & 16,761 video segments          & NVIDIA A40 (48GB) \\
        Action clustering  & SentenceTransformer    & 10,481 unique actions          & CPU-based \\
        Temporal merging   & Rule-based             & 16,761 $\rightarrow$ 14,479 clips & CPU-based \\
        Verification       & Gemini 2.5 Flash Lite  & 14,479 merged segments         & Google API \\
        \bottomrule
    \end{tabular}
\end{table*}

\section{Video Comparison Results}
\label{sec:appendix_video_comparison}

\subsection*{Implementation Details}

Our video comparison framework processed comparisons across 12 \textit{biryani} varieties based on clustered action classes (Table~\ref{tab:clustering_stats}). Since action classes contain multiple video instances, the number of pairwise comparisons grows as $\binom{n}{2}$ where $n$ is the number of clips per action class. Popular action classes like "stirring" (348 instances) and "stirring/mixing rice" (210 instances) (Table~\ref{tab:top_clustering_results})  generated substantially more comparisons than smaller classes.

\begin{table*}[h]
    \centering
    \caption{Top action classes by instance count from clustering results}
    \label{tab:top_clustering_results}
    \begin{tabular}{lc}
        \toprule
        \textbf{Action Class} & \textbf{Instances} \\
        \midrule
        stirring & 348 \\
        stirring/mixing rice & 210 \\
        pouring rice and liquid & 169 \\
        placing/removing pressure cooker lid & 142 \\
        scooping rice and ingredients & 134 \\
        stirring pot contents & 130 \\
        preparing onions & 127 \\
        mixing ingredients in the pot & 125 \\
        serving the \textit{biryani} & 112 \\
        assembling chicken and rice & 107 \\
        stirring/adding chicken & 106 \\
        stirring the mixture & 102 \\
        \bottomrule
    \end{tabular}
\end{table*}

The Proposer stage (Qwen2.5) ran once per action class to generate plausible variations. The Frame Localizer (CLIP with ViT-BigG-14) processed every clip instance within each action class. Both components operated on NVIDIA A40 GPUs with 48GB VRAM, requiring approximately 40 hours each. The Action Differencer used Gemini 2.5 Flash Lite in batch processing mode for final comparisons.

\begin{table*}[h]
    \centering
    \caption{Implementation details for video comparison framework components}
    \label{tab:implementation_details}
    \begin{tabular}{llll}
        \toprule
        \textbf{Component} & \textbf{Model} & \textbf{Processing Scope} & \textbf{Compute Requirements} \\
        \midrule
        Proposer & Qwen2.5 & Once per action class & NVIDIA A40 (48GB), $\sim$40 hours \\
        Frame Localizer & CLIP ViT-BigG-14 & Every clip instance & NVIDIA A40 (48GB), $\sim$40 hours \\
        Action Differencer & Gemini 2.5 Flash Lite & Pairwise comparisons & Batch processing mode through the Gemini API \\
        \bottomrule
    \end{tabular}
\end{table*}

\subsection*{Regional Variation Analysis}

Cross-regional comparisons reveal consistent patterns where certain cooking stages maintain similarity across \textit{biryani} types while others exhibit substantial variation. For each pairwise regional comparison (\textit{Hyderabadi} vs \textit{Kolkata}, \textit{Hyderabadi} vs \textit{Lucknowi}, etc.), fundamental preparation chapters remain consistent while specific execution stages diverge based on cultural techniques.

\begin{figure*}[h]
    \centering
    \includegraphics[width=\textwidth]{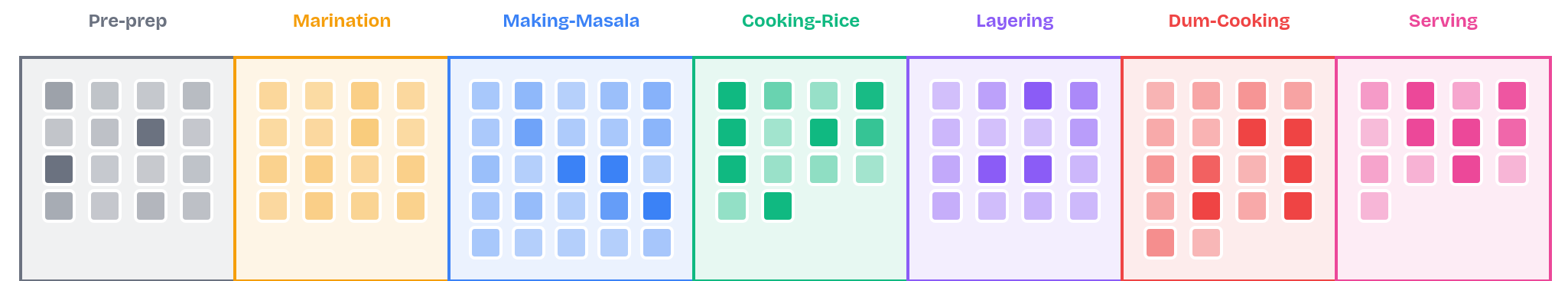}     
    \caption{\textit{Hyderabadi} \textit{biryani} vs \textit{Kolkata biryani} variation visualization. Node opacity indicates the degree of detected procedural differences across cooking stages.}
    \Description{Figure showing a set of chapters \ stages for cooking a biryani, each stage contains actions, the variation of the actions between the Hyderabadi and Lucknowi style of cooking is displayed through the size of the node.}
    \label{fig:regional_comparison}
\end{figure*}

\subsection*{Comparison Statistics}

The framework detected differences in 33.2\% of total comparisons. This percentage represents comparison-level detection: if any proposed difference within a comparison pair was identified, the entire comparison was counted as "difference detected." A comparison was marked as having differences even if only one of multiple proposed variations was found.

If measuring absolute difference detection rather than comparison-level detection, the rate would be approximately 19\%, reflecting the granular nature of individual variation identification within each comparison.

For manual verification accuracy assessment, we use individual difference detection, counting each specific proposed difference separately. For manual verification, we want to know how well our model performed rather than how varied our data is. 

\begin{table}[h!]
    \centering
    \caption{Manual verification accuracy across categories}
    \label{tab:verification}
    \begin{tabular}{lcc}
        \toprule
        \textbf{Category} & \textbf{Correct (\%)} & \textbf{Incorrect (\%)} \\
        \midrule
        Difference detected & 67.5 & 32.5 \\
        No difference       & 45.7 & 54.3 \\
        \bottomrule
    \end{tabular}
\end{table}

\subsection*{Future Improvements}

The framework's limitations suggest specific enhancement directions:

\begin{itemize}
\item \textbf{Enhanced Proposer knowledge}: Deeper understanding of Indian cooking techniques would enable generation of more comprehensive difference categories, particularly when processing large clip volumes per action class.

\item \textbf{Fine-tuned visual encoding}: CLIP's general training may miss fine-grained cooking actions specific to Indian culinary contexts. Increasing retrieved frame counts or specialised model fine-tuning could improve detection granularity.
\end{itemize}

Despite current limitations, the framework successfully captures meaningful procedural differences across regional \textit{biryani} varieties, providing systematic insights into traditional cooking method diversity.

\end{document}

%% file: content/introduction.tex
\section{Introduction}

\begin{figure}[h]
  \centering
  \includegraphics[width=0.95\columnwidth]{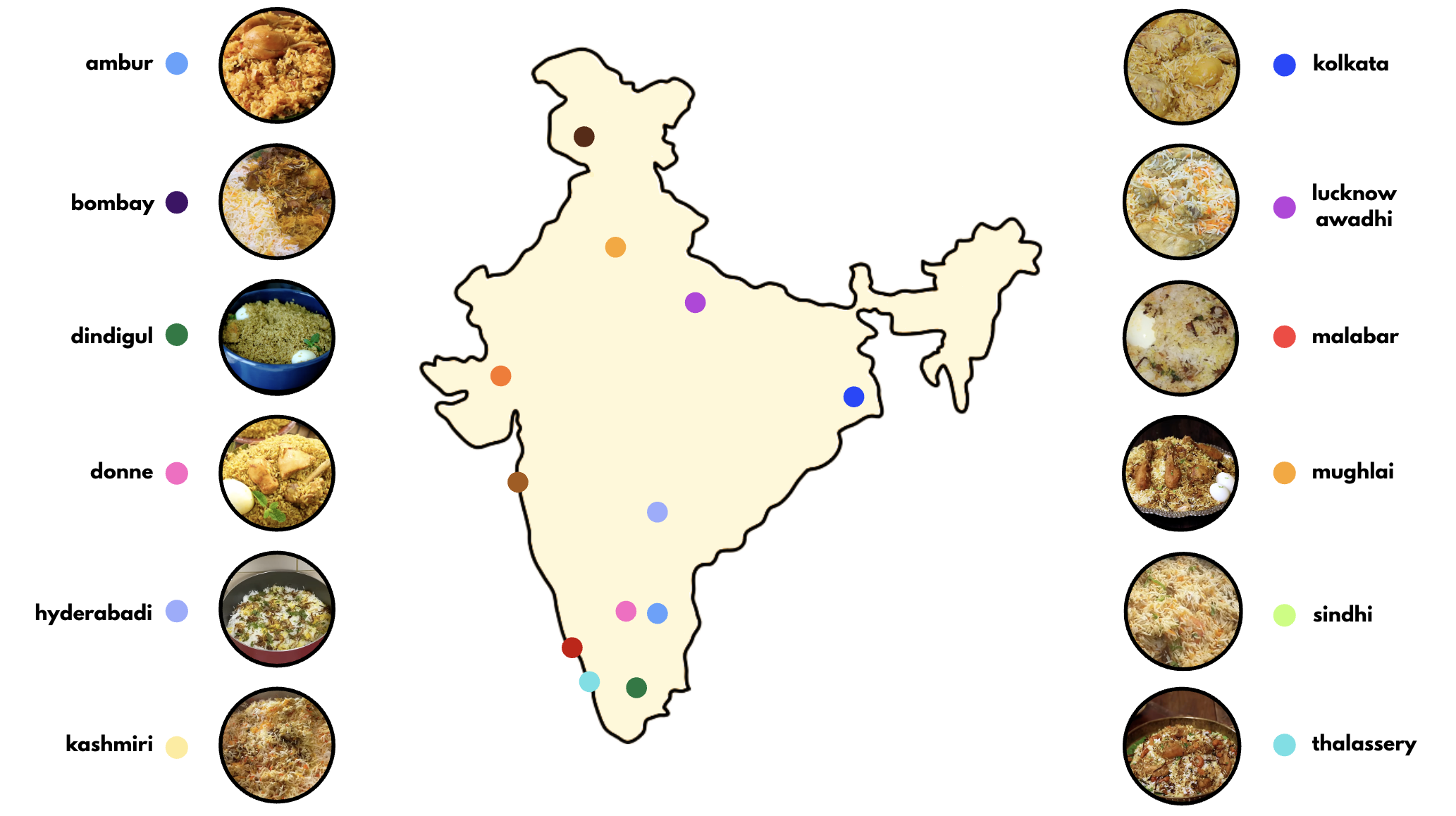}
  
  \caption{Map of India showing 12 regional biryani types -  Ambur, Bombay, Dindigul, Donne, Hyderabadi, Kashmiri, Kolkata, Awadhi, Malabar, Mughlai, Sindhi, and Thalassery. Representative images illustrate differences in preparation, ingredients, and presentation, with all videos sourced from YouTube to capture authentic regional cooking practices.}
  \Description{A map of India with coloured dots marking the origin of 12 regional biryani styles: Ambur, Bombay, Dindigul, Donne, Hyderabadi, Kashmiri, Kolkata, Awadhi, Malabar, Mughlai, Sindhi, and Thalassery. Each type is represented with a small circular food image beside the map, showing differences in rice colour, garnish, and plating.}
  
  \label{fig:biryanimap}
\end{figure}

\Biryani is more than a culinary dish; it is a cultural symbol that embodies the diversity and richness of Indian gastronomy. While its name is shared across the country, its preparation varies widely across regions, shaped by local traditions, availability of ingredients, and individual cooking styles. These differences manifest in flavour and the sequence of preparation steps, the choice of utensils, and the presentation style. With the proliferation of online platforms such as YouTube, this diversity is now documented at scale through cooking videos, providing an invaluable record of culinary practices. However, despite abundant content, the computational tools required to systematically capture and compare fine-grained procedural variations in such videos remain underdeveloped.

\begin{figure*}[t]
	\centering
	\includegraphics[width=0.9\textwidth]{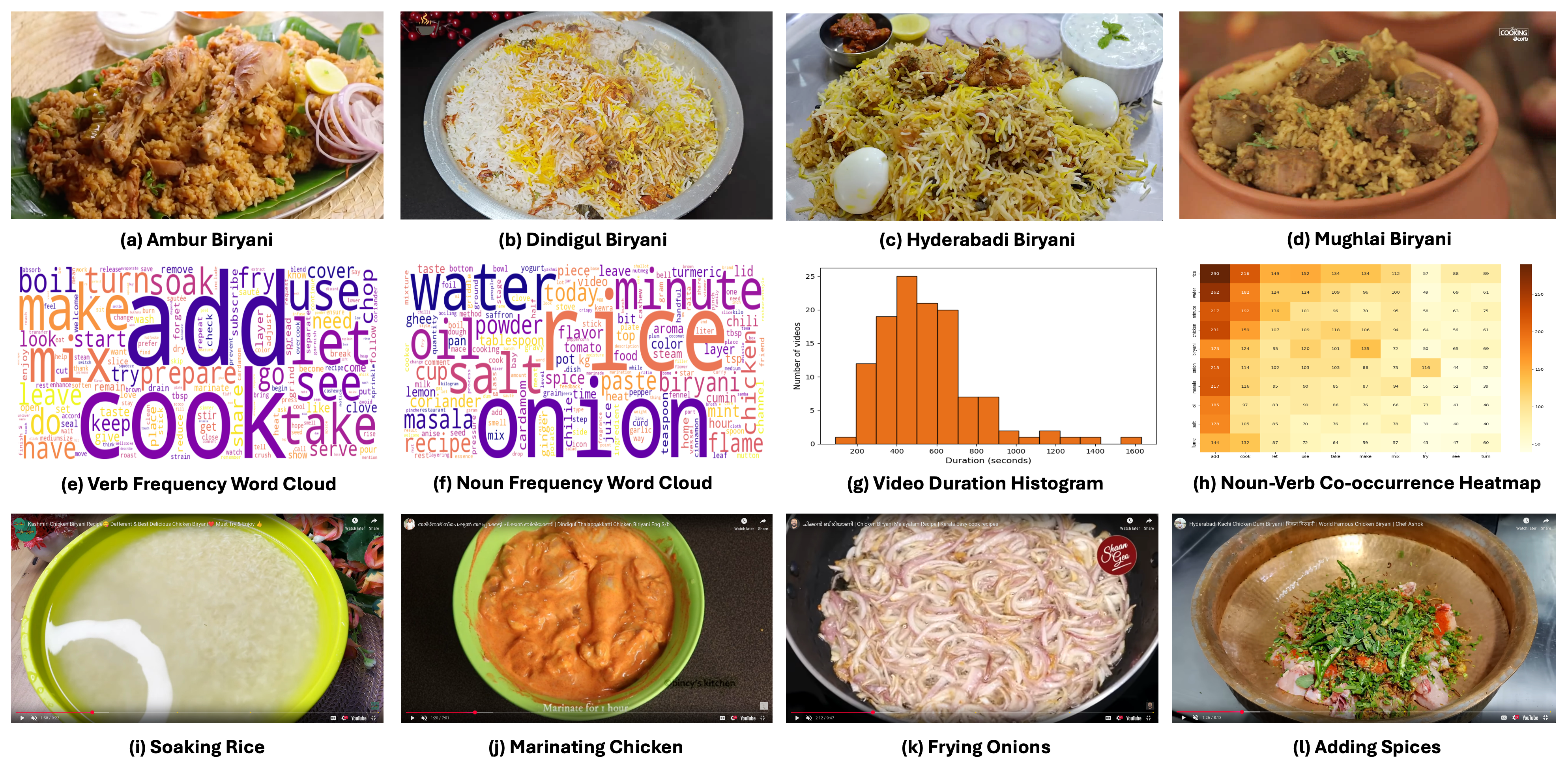}
	\caption{Overview of the Biryani Dataset. Panels (a–d) show representative frames from four of the twelve biryani categories - Ambur, Dindigul, Hyderabadi, and Mughlai - capturing regional diversity in presentation, colour palette, and plating. Panels (e) and (f) present verb and noun frequency word clouds derived from ASR-transcribed and translated speech, revealing common procedural actions and key ingredients. Panel (g) shows the distribution of video durations, with most videos between 5-12 minutes, while panel (h) visualises a noun–verb co-occurrence heatmap, highlighting frequent action–ingredient pairings central to biryani preparation. Panels (i–l) depict canonical procedural steps identified via GPT-4-generated template recipes.}
    \Description{A composite image illustrating visual, textual, and statistical aspects of a biryani video dataset. The top row contains four food images showing different biryani varieties: Ambur Biryani on a banana leaf with sliced onions; Dindigul Biryani in a metal pot with visible rice and spices; Hyderabadi Biryani garnished with boiled eggs; and Mughlai Biryani served in a clay pot. The second row contains two colorful word clouds, one for verbs (large words include “cook,” “add,” “boil,” “use”) and one for nouns (large words include “onion,” “water,” “minute,” “salt”), a histogram showing most videos are between 5 and 12 minutes long, and a heatmap showing how frequently certain nouns and verbs co-occur. The bottom row contains four cooking step images: soaking rice in water, marinating chicken in spices, frying sliced onions in oil, and adding chopped herbs and spices to a pot.}
    \label{fig:dataset_grid}
\end{figure*}

Cooking videos present a unique challenge for computer vision due to their multimodal nature, temporal complexity, and diversity in visual presentation~\cite{whatscookin2015,epickitchens2018,youcook2018,xu2020benchmarkemnlp,hdepic2025,oscar2025}. The same high-level dish can be prepared using markedly different sequences of actions, ingredient combinations, and stylistic elements, often accompanied by narration in different languages or dialects. Indian cooking is known for its multi-step processes and intricate use of spices, and these details are central to understanding the cultural and procedural identity of a recipe~\cite{ktachaya1994,srinivas2011exploring,antani2022evolution}. Conventional video understanding approaches have primarily focused on coarse-grained action recognition or highlight detection, which are insufficient for modelling such nuanced, structured tasks~\cite{whatscookin2015,gao2021learning,nishimura2024recipe,jiao2021survey}.

Over the past two decades, video analysis has evolved from handcrafted feature-based methods~\cite{nanni2017handcrafted,al2020review}, such as Hidden Markov Models~\cite{hmm2001,hmm2002,hmm2018} and Support Vector Machines~\cite{svm2014,svm2020,svm2021}, to deep learning models capable of capturing richer visual patterns from large datasets~\cite{nishani2017computer,sharma2021video,pattichis2023review}. More recently, large vision–language models (VLMs) have emerged as a powerful paradigm, jointly reasoning over visual and textual information to produce semantically meaningful outputs~\cite{zhu2023minigpt,maaz2023video,tang2025video}. These models have demonstrated strong generalisation capabilities in diverse domains, yet their application to structured procedural understanding remains relatively unexplored, particularly in culturally rich contexts. In the context of cooking, and \biryani in particular, VLMs can move beyond recognising individual actions toward modelling the full procedural flow, aligning it with textual recipes, and enabling fine-grained comparisons between variations.

The contributions of this work are as follows:

\begin{itemize}
    \item We introduce the first curated dataset of Indian \biryani preparation videos, annotated with fine-grained temporal segmentation and multimodal labels.
    \item We design a robust VLM-based pipeline for procedural video segmentation, multimodal alignment, and question-answer generation.
    \item We propose a novel video comparison framework for analysing subtle procedural differences across regional \biryani variants.
    \item We provide quantitative benchmarks and qualitative analyses of the performance of current VLMs on culturally rich procedural video understanding tasks.
\end{itemize}
    
The remainder of the paper is organised as follows. Section 2 describes the dataset curation process, Section 3 details the video segmentation framework, Section 4 presents the multimodal alignment methodology, Section 5 outlines the QA dataset generation and benchmarking experiments, Section 6 introduces the video comparison framework, and Section 7 discusses potential applications and concludes.

%% file: content/dataset_and_viz.tex
\section{Biryani Dataset}

We want to study how different videos curated for the same purpose (in this case cooking \textit{biryani}) differs or compares. We start with creating a dataset of publicly available \biryani cooking videos. 
We curate a dataset of 120 cooking videos focused on \biryani preparation, sourced from YouTube. The dataset spans 12 distinct types of \biryani (Ambur, Bombay, Dindigul, Donne, Hyderabadi, Kashmiri, Kolkata, Awadhi, Malabar, Mughlai, Sindhi, and Thalassery).
For each category, we collect 10 distinct videos per category, as shown with representative frames in Figure~\ref{fig:dataset_grid} (a-d), illustrating the diversity in presentation, colour palette, and plating traditions across regions.

Videos were chosen for their culinary popularity and the availability of high-quality recordings. To maximise utility for the downstream tasks, we prioritized videos featuring clear audio, spoken narration of cooking steps, complete visual coverage of the preparation process, and a range of durations. Given the pan-Indian diversity of the selected \biryani types, the dataset exhibits substantial variation in language, cooking techniques, narration styles, and cinematographic choices such as camera angles and editing styles.

We first extract audio from each video and perform automatic speech recognition (ASR) using WhisperX~\cite{bain2023whisperx} and Whisper-Large \cite{radford2023robust}. All transcripts are translated into English (using GPT-4 ~\cite{achiam2023gpt}) to standardise linguistic representation across the dataset. We then use part-of-speech tagging with spaCy~\cite{honnibal2020spacy} to extract nouns, verbs, and adjectives from the transcripts, producing frequency-based visualisations such as word clouds. Figures~\ref{fig:dataset_grid} (e, f) show examples of these visualisations, where high-frequency verbs (e.g., “add”, “cook”) and nouns (e.g., “rice”, “onion”) capture the procedural and ingredient focus of \biryani preparation. Additional analyses, such as the duration histogram in 
Figure~\ref{fig:dataset_grid} (g), reveal that most videos fall within a 5–12 minute range, while the noun–verb co-occurrence heatmap in Figure~\ref{fig:dataset_grid} (h) highlights common action–object pairings that define core cooking steps.

To enable fine-grained analysis (such as step-level captioning or instruction grounding), we segment each cooking video into meaningful procedural units. We generate canonical template recipes for each \biryani type using GPT-4 \cite{achiam2023gpt}, which provided structured reference sequences of cooking steps. These generated templates served as a standardized framework for identifying procedural steps across diverse video formats, rather than acting as an authentic recipe ground-truth. Manual verification ensured the consistency and usability of this framework for temporal segmentation. An additional ``Miscellaneous/Intro/Outro'' class is used in each template to account for non-cooking content commonly present in YouTube videos, such as greetings, personal anecdotes, promotional messages, or outros, ensuring that such segments are meaningfully grouped and excluded from step-level misalignment. Figure~\ref{fig:dataset_grid} (i–l) depicts canonical procedural frames extracted from videos, including soaking rice, marinating chicken, frying onions, and adding spices—steps that recur across multiple \biryani variants despite regional differences.

%% file: content/video_seg.tex
\section{Video Segmentation}

\begin{figure*}[t]
    \centering
    \includegraphics[width=0.95\textwidth]{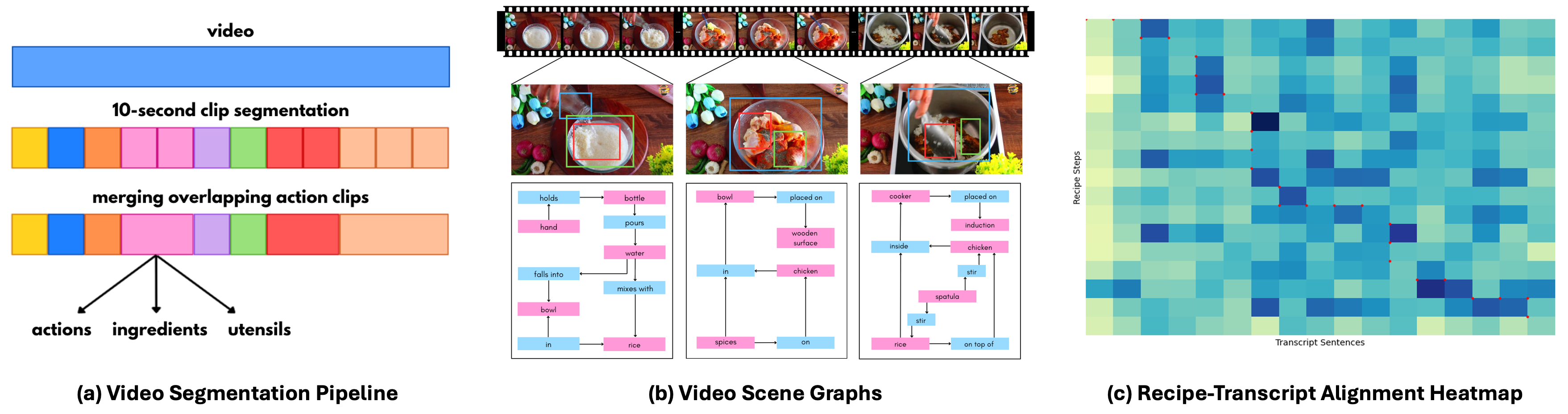}
    \caption{Overview of the multimodal video segmentation and alignment pipeline. Panel (a) shows the 10-second clip segmentation of \biryani cooking videos, where each segment is processed by InternVL-14B to extract visually grounded annotations of actions, ingredients, and utensils. Consecutive segments containing the same action are merged to form continuous spans, improving temporal coherence. Panel (b) presents example video scene graphs depicting detected entities and their relationships.  displays an alignment heatmap between canonical recipe steps (vertical) and transcript sentences (horizontal), where colour intensity indicates semantic similarity and the red path represents the optimal sequence alignment computed via Dynamic Time Warping.}
    \label{fig:video_seg_pipeline}
    \Description{The figure has three panels. 
Panel (a) shows the video segmentation pipeline, where \biryani cooking videos are divided into 10-second clips, each annotated by InternVL-14B with actions, ingredients, and utensils. Consecutive clips containing the same canonicalised action are merged into continuous spans to improve temporal coherence. 
Panel (b) presents sample video scene graphs, depicting detected ingredients and utensils linked by their associated actions, shown alongside representative video frames. 
Panel (c) is a heatmap illustrating alignment between canonical recipe steps and transcript sentences. Colour intensity indicates semantic similarity, and a red path traces the optimal alignment computed with Dynamic Time Warping.}

\end{figure*}


We use InternVL-14B~\cite{zhu2025internvl3}, a state-of-the-art Vision-Language Model (VLM), to process each segment. As shown in Fig.~\ref{fig:video_seg_pipeline}, the model is prompted to extract three key categories of information: (a) Ingredients, (b) Utensils (Objects), and (c) Actions (verbs). Significantly, the model relies solely on visual content (sampled video frames) and does not access audio or transcripts, ensuring that annotations are grounded purely in visual evidence.\footnote{All the prompts used in this paper are available in Appendix~\ref{sec:appendix_prompts_used}.} Since cooking actions often span more than one 10-second interval, the same canonicalised action label can appear in consecutive segments. To improve temporal coherence, we merge timestamps for such repeated actions within a video into a single continuous span, while ensuring unrelated actions in adjacent segments remain separate. This reduces unnecessary fragmentation and yields longer, coherent action-level sequences without merging distinct activities.
Direct application of InternVL-14B across thousands of segments yields a detailed mapping of ingredients, utensils, and actions over time. However, action descriptions often vary lexically despite being semantically identical (e.g., “stirring rice” vs. “stirring rice and water with a wooden spoon”). To address this, each action phrase is embedded using the all-MiniLM-L6-v2 SentenceTransformer model \cite{reimers-2019-sentence-bert} and clustered via agglomerative clustering with average linkage and a cosine distance threshold of 0.3, merging clusters until no pair falls below this threshold. A single representative phrase from each cluster serves as the canonical action label, improving label consistency and enabling robust querying, statistical analysis, and downstream tasks such as recipe step generation and video retrieval.

Although InternVL-14B produced high-quality visual annotations, we introduced an automated verification step using  Gemini-2.5-flash-lite \cite{gemini25} to ensure each labelled action was visibly present in its corresponding segment. This lightweight VLM was queried with deterministic yes/no prompts over sampled video frames, enabling reliable validation for downstream tasks such as step-wise recipe alignment and skill-specific retrieval. \footnote{The complete verification workflow is provided in Appendix~\ref{sec:appendix_video_segmentation}.}. We verified 14,470 video–action segments across all \biryani types, with 11,295 (78.05\%) labelled as correct and 3,175 (21.95\%) as incorrect, thereby increasing confidence in the dataset’s action labels.

\subsection{Results}
The initial action detection stage produced a highly granular label space, with 10,481 unique action classes.  
After applying the action clustering process, this number was reduced to 2,187 canonicalised action classes, greatly improving consistency in labelling.

Similarly, the temporal merging process significantly reduced fragmentation in the video segmentation.  
Across all videos, the number of timestamped clips decreased from 16,761 before merging to 14,479 after merging, representing a 13.6\% reduction in segment count while preserving full action coverage.

\subsection{Multimodal Alignment: Video, Audio, and Recipe Texts}

To build a unified understanding of each \biryani cooking video, we align three modalities: WhisperX transcripts (temporally ordered narration), InternVL visual segment descriptions (ingredients, utensils, and actions for every 10-second chunk), and manually curated canonical recipes (standard steps and titles per \biryani type). As shown in Fig~\ref{fig:video_seg_pipeline}, alignment begins with coarse filtering, where lowercased and tokenised segment metadata keywords (from detected ingredients/utensils) are matched against transcript lines and recipe steps to eliminate irrelevant pairs. Remaining candidates undergo fine-grained alignment: transcript sentences and recipe steps are embedded with a SentenceTransformer \cite{reimers-2019-sentence-bert}, and Dynamic Time Warping (DTW) over cosine distances preserves sequential structure while tolerating omissions, insertions, or reordering—handling deviations from ideal diagonal mappings caused by narration order, granularity mismatches, or pacing differences. For segments passing coarse filtering, we further embed InternVL-extracted actions and recipe steps using BGE \cite{bge_embedding}, compute cosine similarities, assign each chunk to its most semantically relevant recipe step, and rank segments per step with confidence scores. This multimodal alignment enables recipe-aware search, visualisation, and retrieval across heterogeneous time scales and structures.


%% file: content/video_comparison.tex
\section{Video Comparison}

We aim to understand different \biryani recipes by comparing their cooking processes. By comparing the cooking process for different types of \textit{biryani}, we can identify common patterns and variations in the cooking methods, ingredients, and techniques used. This can help us understand the unique characteristics of each \biryani recipe and how they differ.

To compare the cooking processes, including ingredients, methods, actions, etc., different \biryani varieties (for example, \textit{Hyderabadi biryani} vs \textit{Lucknowi biryani}), we developed a video comparison framework adapted from the VidDiff method \cite{burgess2025video} that identifies and visualises the differences in cooking actions, ingredients, and techniques. This framework is designed to analyse the cooking videos in our dataset, allowing users to understand how different \biryani recipes vary in terms of their preparation methods.

\begin{figure*}
	\includegraphics[width=0.95\textwidth]{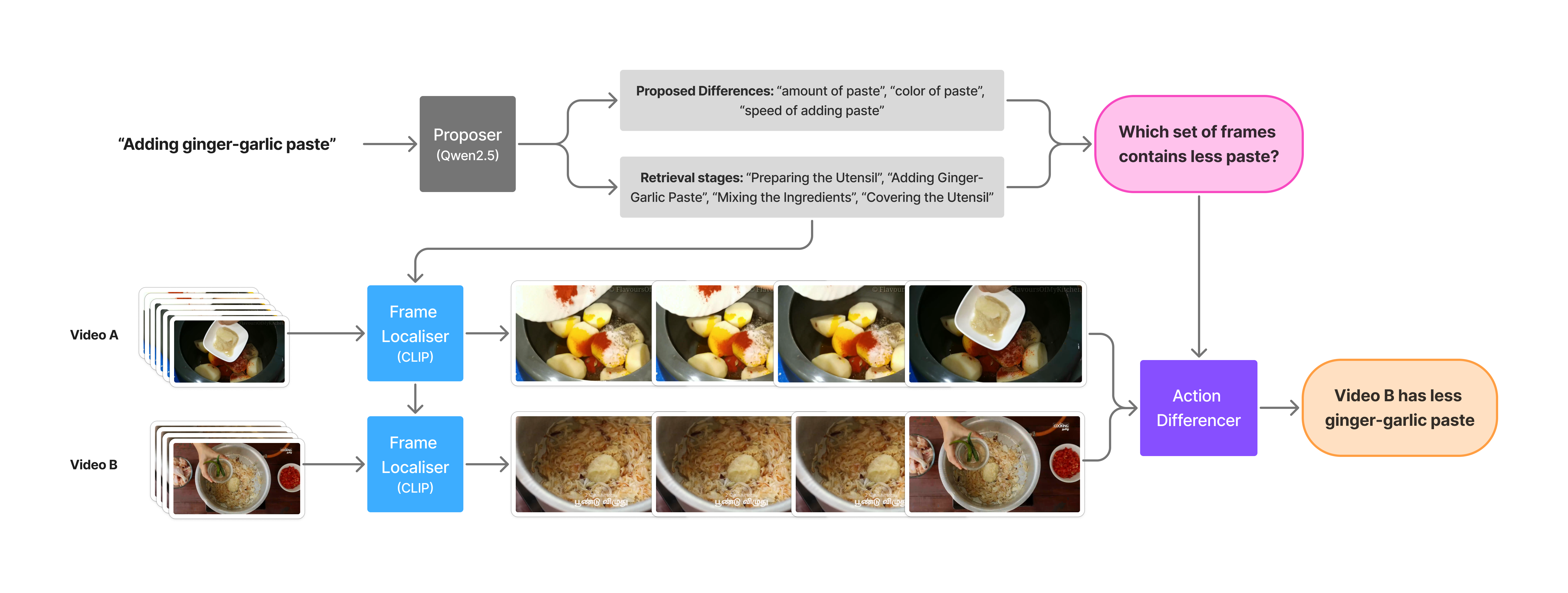}
	\caption{Overview of the video comparison framework for \biryani recipes. The framework operates through three sequential stages: Proposer (Qwen2-VL) generates plausible variations for each action, Frame Localiser (CLIP) identifies relevant frames, and Action Differencer compares frame pairs to detect differences. This example demonstrates analysis of "Adding ginger-garlic paste," identifying that Video B uses less paste than Video A.}
	\Description{Figure shows the video comparison framework for \biryani{} recipes. The framework consists of three main stages: Proposer, Frame Retriever, and Action Differentiating. The Proposer stage generates plausible variations for each action class, the Frame Retriever stage retrieves the temporal localisation of sub-actions from cooking videos, and the Action Differentiating stage analyses and visualises differences between two cooking video segments.}
	\label{fig:comparison_pipline}
\end{figure*}

To compare the cooking processes across different \biryani recipes, we adapted the VidDiff framework \cite{burgess2025video} to our specific use case. The framework consists of three main stages:

\paragraph{Proposer:} This stage generates plausible variations for each action class. For each action class, we prompt an LLM to generate plausible ways the action might vary. We also take an action and break it down into sub-actions. Finally, we link the differences to the sub-actions. The LLM is prompted to generate 2-3 variations in the cooking actions that are visually significant and would affect the final frames, and also prompted to generate 2-4 sub-action stages for each action class. The LLM then creates explicit mappings between variations and sub-actions. These mappings specify which differences would be most visually detectable during specific sub-action stages. We employ Qwen2.5 \cite{qwen2.5}.

\paragraph{Frame Retriever:} This stage retrieves temporal localisation of sub-actions from cooking videos using CLIP \cite{reimers-2019-sentence-bert}. We embed textual retrieval strings and video frames into a shared semantic space, then compute cosine similarity scores to identify the top-k (k=2) frames that best match each sub-action. This focuses on peak similarity moments where sub-actions are most visually apparent, using ViT-BigG-14 (Open-CLIP) \cite{ilharco_gabriel_2021_5143773}.

\paragraph{Action Differentiating:} In this final stage, we analyse and visualise the differences between two cooking video segments (segmented by action) using the last stage's localised frames. For each pair of corresponding sub-action segments identified in the previous stage, we pose a multiple-choice question (which were generated from the multiple differences we got from the proposer stage) to a VLM, which determines whether each difference is present in \textit{Video A} or \textit{Video B} or \textit{It's unsure}. We transform our recipe comparison task into a multiple-choice question for the VLM. The VLM is then used to determine which video shows more of the proposed difference, providing a detailed explanation of the observed differences. This allows us to visualise and understand how the cooking processes differ between the two \biryani recipes. We employ Gemini-2.5-flash-lite \cite{gemini25}.

\subsection{Results}

\begin{figure*}[!t]
	\centering
	\includegraphics[width= 0.95\textwidth]{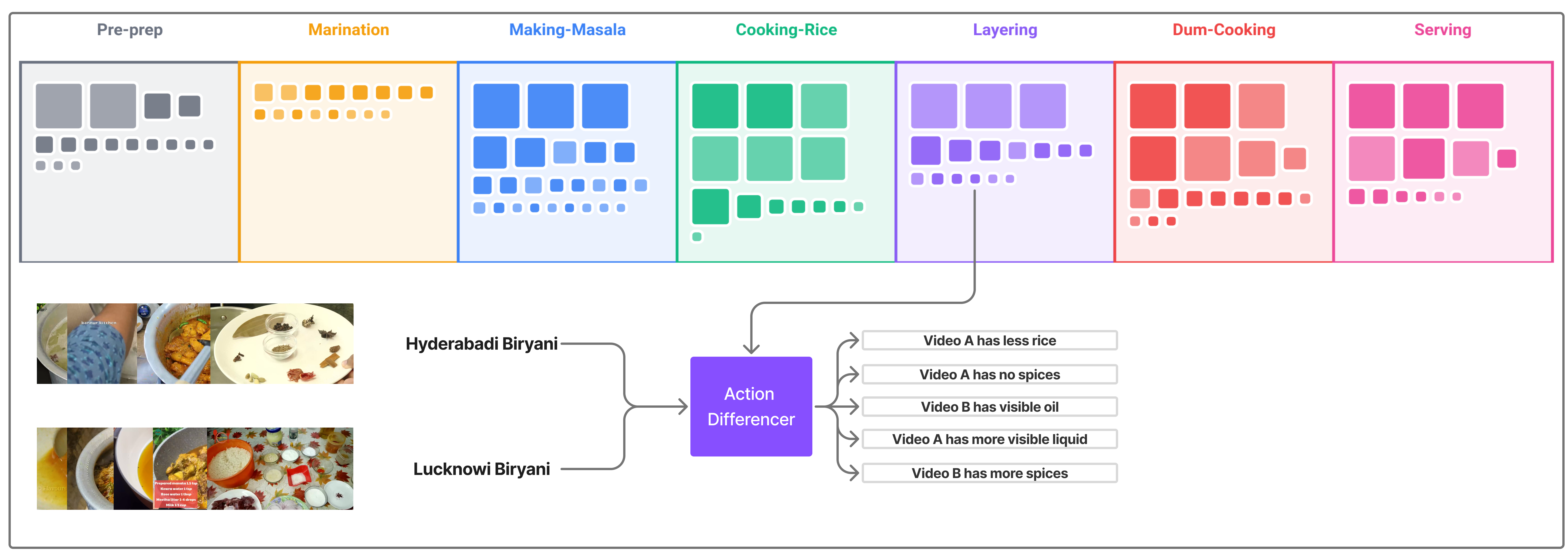}
	\caption{Visualisation of cooking process variations between \textit{Hyderabadi} and \textit{Lucknowi} \biryani across several cooking stages. Each coloured section represents a major cooking stage, with individual squares showing specific actions. The opacity of the square is proportional to the degree of variation detected between the two \biryani styles, where larger squares indicate more significant differences. }
	\Description{A timeline visualisation showing seven cooking stages (Pre-prep, Marination, Making-Masala, Cooking-Rice, Layering, Dum-Cooking, Serving) with squares of varying opacity representing the degree of variation detected between Hyderabadi and Lucknowi biryani preparation methods.}
	\label{fig:comparison_result}
\end{figure*}

Our video comparison framework identified meaningful differences across \biryani varieties. Figure~\ref{fig:comparison_result} shows that certain cooking stages exhibit minimal variation between \textit{Hyderabadi} and \textit{Lucknowi} \biryani, while others display substantial differences.

\begin{table}[h!]
	\centering
	\caption{Distribution of comparison results across randomly paired video segments from 12 \biryani varieties}
	\label{tab:comparison_results}
	\begin{tabular}{lc}
		\toprule
		\textbf{Outcome}         & \textbf{Percentage of Comparison} \\
		\midrule
		Difference detected      & 33.2\%                            \\
		No detectable difference & 66.8\%                            \\
		\bottomrule
	\end{tabular}
\end{table}

The framework detected differences in 33.2\% of action comparisons. This proportion indicates that while biryani varieties share core cooking procedures, they exhibit distinct variations in execution methods. The detection rate aligns with expectations for regional culinary variants, where fundamental processes remain consistent but specific techniques diverge based on cultural and regional influences.

Further details, visualisations, and discussion of these results are provided in Appendix~\ref{sec:appendix_video_comparison}.

To validate the accuracy of the framework, 2000 randomly sampled comparisons were verified by a group of 4–5 independent annotators, with each annotator reviewing a subset of the samples. The verification focused on confirming model-proposed differences rather than performing exhaustive difference detection, which would scale exponentially and is not practically feasible. Table~\ref{tab:verification} shows accuracy rates across different comparison categories.

\begin{table}[h!]
	\centering
	\caption{Manual verification accuracy across categories}
	\label{tab:verification}
	\begin{tabular}{lcc}
		\toprule
		\textbf{Category}   & \textbf{Correct} & \textbf{Incorrect} \\
		\midrule
		Difference detected & 67.5\%           & 32.5\%             \\
		No difference       & 45.7\%           & 54.3\%             \\
		\bottomrule
	\end{tabular}
\end{table}

The verification results reveal systematic challenges in the model's performance. The framework achieved 67.5\% accuracy for detected differences, indicating reliable identification of actual procedural variations. However, accuracy drops to 45.7\% for "no difference" classifications, suggesting the model misses subtle but meaningful variations that human annotators can detect. This performance gap likely stems from the model's limited exposure to Indian cooking contexts during training, resulting in conservative judgments when analysing culturally specific culinary techniques. Additionally, the model occasionally generates false differences or misattributes variations between video clips, highlighting areas for future improvement.

Despite these limitations, the framework successfully captures meaningful procedural differences across regional biryani varieties, providing valuable insights into how traditional cooking methods vary while maintaining cultural authenticity.

%% file: content/qa_gen.tex
\begin{figure*}[t]
    \centering
    \includegraphics[width=0.99\textwidth]{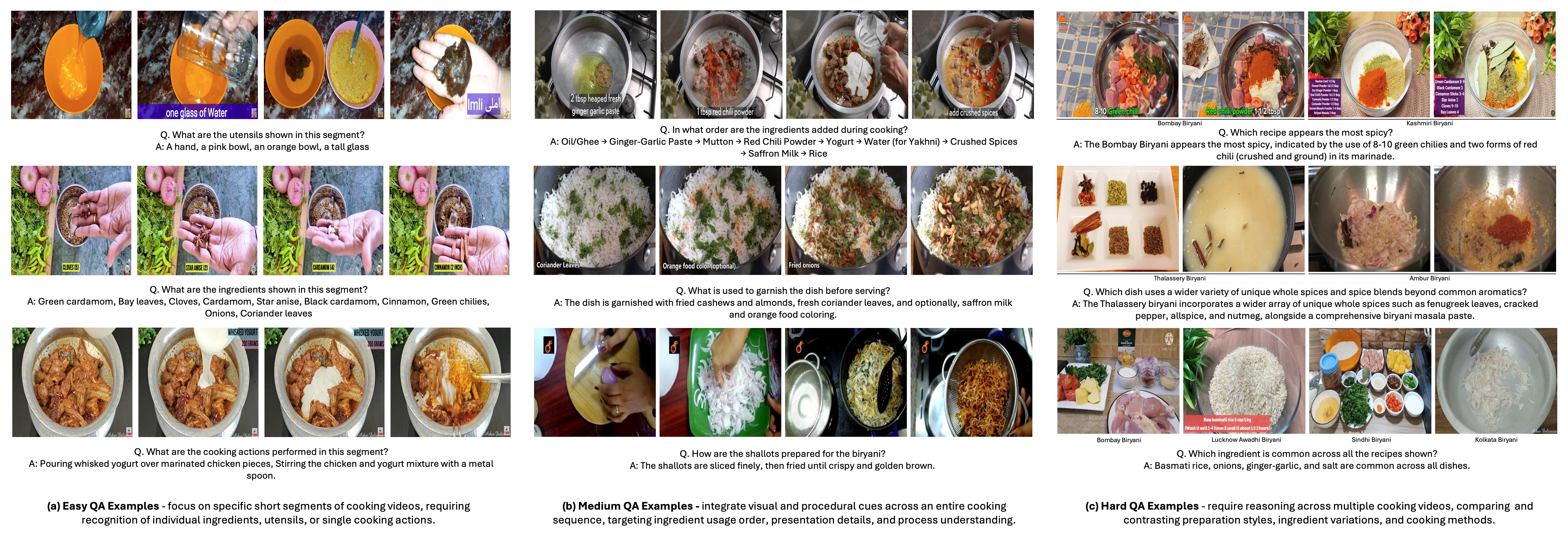}
    
    \caption{Example QA pairs from the biryani video QA dataset, covering easy, medium, and hard difficulty tiers. Questions were generated via a multi-stage pipeline (temporal segmentation, captioning, summary synthesis, LLM prompting, and human curation). Easy QAs are segment-level recognition tasks, medium QAs require whole-video temporal and procedural understanding, and hard QAs demand multi-video reasoning and comparison. These examples illustrate the dataset’s progression from simple perception to complex reasoning.}
     
    \Description{
    A multi-row image panel showing question-answer pairs for cooking video understanding. 
    \textbf{(a) Easy QA} contains examples such as identifying utensils (e.g., pink bowl, orange bowl, tall glass), listing visible ingredients (e.g., green cardamom, bay leaves, cinnamon), and describing cooking actions (e.g., pouring yogurt over chicken). Each question is paired with annotated video frames.
    \textbf{(b) Medium QA} presents examples requiring sequential understanding, such as the order of adding ingredients during cooking, garnishing steps, and how shallots are prepared for biryani. The images show consecutive cooking frames with ordered ingredient use.
    \textbf{(c) Hard QA} involves reasoning across multiple cooking videos to answer questions like which recipe appears the most spicy, which uses a wider variety of spices, and which ingredient is common across all shown recipes. These examples feature comparison across frames from different biryani styles (e.g., Bombay, Thalassery, Awadhi, Lucknowi, Kolkata).}
    
    \label{fig:qaexamples}
\end{figure*}

\section{Video Question Answering}

Video Question Answering (VQA) is a key benchmark for evaluating comprehensive scene understanding \cite{xu2017video,Jang_2017_CVPR,shang2021video,yang2022avqa,Xiao_2021_CVPR}. Unlike static image tasks, it requires joint reasoning over spatial (object and scene layout), temporal (event ordering, procedural flow), and causal (why actions occur) aspects within and across videos. This capability moves AI/ML systems beyond isolated recognition toward context-aware reasoning in dynamic settings.

In cooking, such reasoning is essential: `What ingredient was added before the onions?' demands temporal ordering; `Why was the heat reduced after adding milk?' requires causal inference; and `Which recipe uses more spices?' involves multi-video comparison. By spanning easy, medium, and hard difficulty tiers, our dataset targets this spectrum—from basic perceptual recognition to complex cross-video reasoning—making it both a challenge for current VLMs and a step toward more general-purpose, reasoning-capable AI.

We construct the dataset using a multi-stage pipeline of temporal segmentation, automated visual description, language model prompting, and manual curation. Difficulty tiers are defined as Easy (single short segment), Medium (entire video comprehension), and Hard (multi-video reasoning). Each video is temporally segmented to capture localised cooking events, with InternVL3-14B \cite{zhu2025internvl3} producing natural language descriptions of ingredients, utensils, and preparation steps. Gemini-2.0-Flash then integrates these segment-level captions \cite{team2025gemini} into coherent, visually detailed, step-by-step recipe narratives that comprehensively represent the entire cooking process.

\subsection{QA Generation}

\paragraph{Easy QA Generation}
For easy QA pairs, we focus on individual segments. We randomly sample up to three 10-second segments for each video to generate QA pairs, balancing diversity and computational efficiency. We prompt Llama-3-8B-Instruct \cite{grattafiori2024llama} to systematically extract three categories of information from each selected segment: 
(a) ingredients shown (b) utensils used (c) cooking actions performed

To ensure high data quality, we manually review the generated QA pairs for each video, selecting the two most informative and unambiguous examples \footnote{Examples of QA pairs for each difficulty tier are provided in Appendix~\ref{sec:appendix_qa_examples}.}. This curation step filters out incomplete, repetitive, or low-detail responses, yielding a robust set of easy, segment-grounded QA pairs. 

\paragraph{Medium QA Generation}
For medium-level QA generation, the goal is to assess the model’s comprehension of the entire cooking process in each video, requiring integration of visual and procedural cues across the full temporal span. In contrast to the short-segment focus of easy QA pairs, these questions target broader aspects such as ingredient usage, temporal ordering of key steps, and presentation details. Video summaries are combined with aligned audio transcripts to enable this, providing a rich multimodal textual context that captures visual observations and spoken instructions. Using this input, we prompt Gemini-2.0-Flash \cite{team2025gemini} to produce a high-level summary and multiple QA pairs, guided by carefully designed question templates tailored to cooking scenarios. These templates emphasize visual elements (e.g., primary ingredients, garnishes, spices), temporal understanding (e.g., sequence of actions, cooking durations, preparation time), and utensil or process details (e.g., vessel type, marination or frying steps), while allowing the model to generate additional contextually relevant questions beyond the provided templates.

\paragraph{Hard QA Generation}

For the most challenging QA tier, we evaluate a model’s ability to reason across multiple cooking videos, requiring deeper comparative understanding of recipes, cooking styles, and ingredient choices. We first create multimodal summaries of individual videos by combining detailed frame-wise visual descriptions with complete audio transcripts, capturing both rich visual details (ingredients, techniques, utensils, textures, plating) and spoken instructions (quantities, tips, emphases).

We generate hard QA pairs from these summaries by sampling combinations of 2, 3, 4, and 5 videos from the 120-video pool and instructing Gemini-2.5-Flash \cite{gemini25} to analyse their combined content. The model compares, contrasts, and synthesizes details—such as ingredients, cooking methods, spice levels, preparation sequences, and presentation styles—to formulate high-level, reasoning-intensive QAs that require integrating information from multiple sources.

\paragraph{\bf Dataset Statistics}

Our QA generation pipeline produces 240 easy, 1,357 medium, and 486 hard question–answer (QA) pairs. The hard QA set is further subdivided based on the number of videos required for reasoning: \texttt{hardqa2} (146), \texttt{hardqa3} (171), \texttt{hardqa4} (82), and \texttt{hardqa5} (87). The dataset is evenly split into training and test sets to support model development and evaluation, ensuring balanced representation across all difficulty levels and subsets.

\begin{figure}[h]
    \centering
    \includegraphics[width=0.5\textwidth]{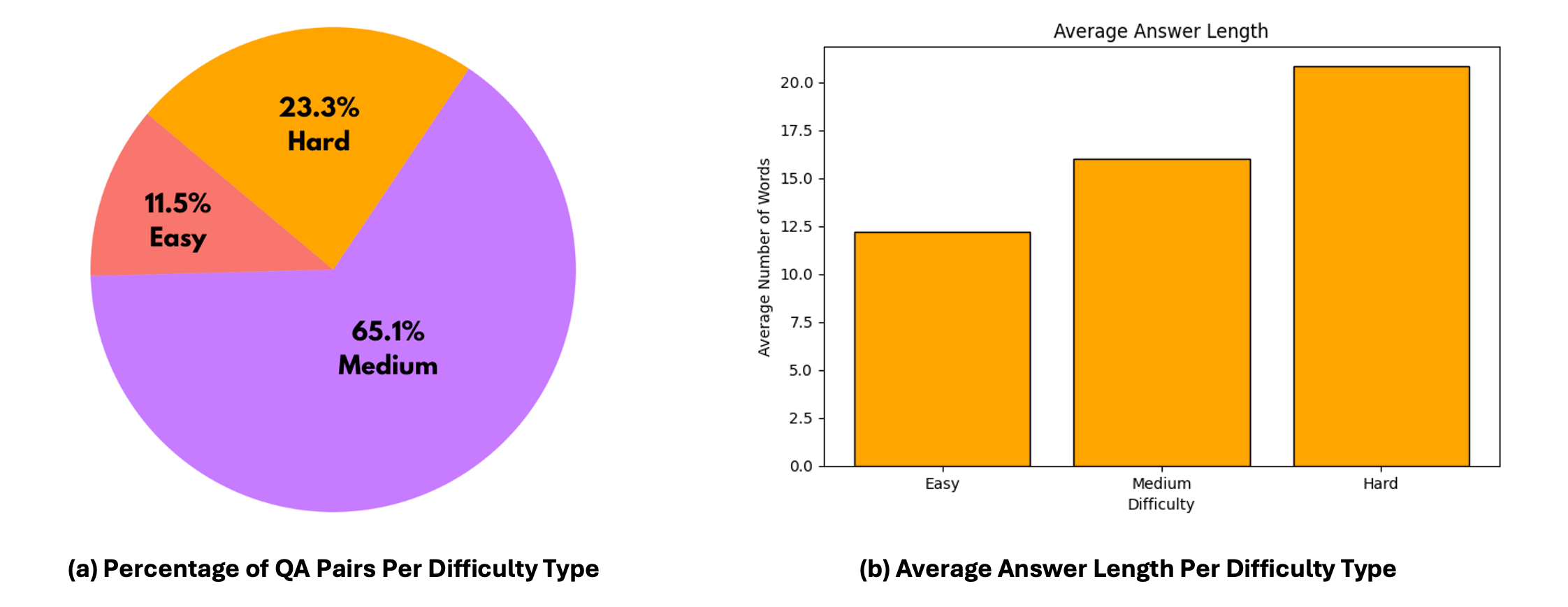}
    
    \caption{Statistics of the biryani video QA dataset. (a) Distribution of question–answer pairs across difficulty levels. (b) Average answer length per difficulty type, showing a clear upward trend with complexity.} 
    \Description{
    The left chart is a pie chart showing the proportion of QA pairs by difficulty: 65.1\% Medium (purple), 23.3\% Hard (yellow-orange), and 11.5\% Easy (pink). 
    The right chart is a bar plot showing average answer length by difficulty: Easy $\approx$ 12 words, Medium $\approx$ 16 words, Hard $\approx$ 20 words, with all bars coloured orange. 
    The figure illustrates that harder questions not only occur less frequently but also tend to have longer answers.}
    
    \label{fig:qastats}
\end{figure}

Figure \ref{fig:qastats} summarizes these statistics. Panel (a) shows the percentage distribution of QA pairs by difficulty, where medium questions dominate (65.1\%), followed by hard (23.3\%) and easy (11.5\%). Panel (b) presents the average answer length for each difficulty type. As expected, harder questions tend to require longer answers, with an average of over 20 words for hard items compared to around 12 words for easy ones. This trend reflects the increased complexity and reasoning demands of higher difficulty levels.

\subsection{Results}


    We benchmark existing video–language models (VLMs) on our QA dataset using both zero-shot and fine-tuned settings. Five open-source VLMs are evaluated in zero-shot mode — InternVL3-8B (internvl3) \cite{zhu2025internvl3}, Qwen2-VL-7B-Instruct (qwen2vl) \cite{Qwen2VL}, llava-v1.6-mistral-7b-hf (llavanext) \cite{liu2024improved}, llava-onevision-qwen2-7b-ov-hf (llava \allowbreak ov) \cite{li2024llava}, and VideoLLaMA3-7B-Image (videollama) \cite{zhang2025videollama} — and we fine-tune Llama-3.2-11B-Vision-Instruct (llama3ft) \cite{grattafiori2024llama} on our dataset with type-specific prompts and frame-sampled inputs to measure domain adaptation gains.

We report standard QA metrics - BLEU, ROUGE-L, and BERTScore - to capture lexical and semantic similarity, but true evaluation lies in the dataset’s tier design. The medium and hard tiers deliberately require temporal, procedural, and cross-recipe reasoning, making the tier structure a stronger indicator of reasoning depth than raw metric scores.k

Across all metrics, the fine-tuned Llama-3.2 outperforms zero-shot baselines, with the most significant gains on medium and hard questions. Improvements are most pronounced in BERTScore, indicating stronger semantic alignment in addition to lexical accuracy. Some zero-shot models (e.g., Qwen2-VL, InternVL3) perform competitively in certain tiers, but none match the fine-tuned model’s consistency.

For the hard QA tier, we further break down results into \texttt{hard2} – \texttt{hard5}, corresponding to the number of videos required for reasoning. Tables \ref{tab:overallmetrics} and \ref{tab:hardmetrics} present full results. Performance generally declines with more videos, reflecting the difficulty of multi-video reasoning.

\begin{table}[h]
\centering
\footnotesize

\caption{Overall QA performance of VLMs on the QA dataset across easy, medium, and hard difficulty tiers. Best results for each metric–tier combination are highlighted in bold.}

\label{tab:overallmetrics}
\begin{tabularx}{\linewidth}{l l X X X }
\toprule
VLM & Metric & easy & medium & hard \\
\midrule
\multirow{3}{*}{internvl3} & BLEU & 0.0294 & 0.0291 & 0.0395 \\
 & ROUGE-L & 0.2184 & 0.1732 & 0.2457 \\
 & BERTScore & 0.1663 & 0.1628 & 0.2683 \\
\midrule
\multirow{3}{*}{qwen2vl} & BLEU & 0.0314 & 0.0209 & 0.0609 \\
 & ROUGE-L & 0.1914 & 0.1189 & 0.3201 \\
 & BERTScore & 0.1298 & -0.0747 & 0.3022 \\
\midrule
\multirow{3}{*}{llavanext} & BLEU & 0.0128 & 0.0216 & 0.0150 \\
 & ROUGE-L & 0.1319 & 0.1367 & 0.1911 \\
 & BERTScore & -0.1732 & 0.0465 & 0.0984 \\
\midrule
\multirow{3}{*}{llavaov} & BLEU & 0.0038 & 0.0278 & 0.0246 \\
 & ROUGE-L & 0.0408 & 0.1383 & 0.1386 \\
 & BERTScore & -0.2586 & 0.0377 & -0.0073 \\
\midrule
\multirow{3}{*}{videollama} & BLEU & 0.0194 & 0.0787 & 0.0502 \\
 & ROUGE-L & 0.1883 & 0.2713 & 0.2650 \\
 & BERTScore & 0.0897 & 0.3071 & 0.2445 \\
\midrule
\multirow{3}{*}{llama3ft} & BLEU & \textbf{0.0472} & \textbf{0.1683} & \textbf{0.1140} \\
 & ROUGE-L & \textbf{0.2689} & \textbf{0.4214} & \textbf{0.4072} \\
 & BERTScore & \textbf{0.2660} & \textbf{0.4869} & \textbf{0.4526} \\
\bottomrule
\end{tabularx}
\end{table}

\begin{table}[h]
\centering
\footnotesize

\caption{Hard-tier breakdown showing VLM performance on subsets \texttt{hard2}, \texttt{hard3}, \texttt{hard4}, and \texttt{hard5}, corresponding to the number of videos required for reasoning.}

\label{tab:hardmetrics}
\begin{tabularx}{\linewidth}{l l X X X X }
\toprule
VLM & Metric & hard2 & hard3 & hard4 & hard5 \\
\midrule
\multirow{3}{*}{internvl3} & BLEU & 0.0432 & 0.0405 & 0.0386 & 0.0322 \\
 & ROUGE-L & 0.2624 & 0.2510 & 0.2444 & 0.2087 \\
 & BERTScore & 0.2882 & 0.2756 & 0.2532 & 0.2347 \\
\midrule
\multirow{3}{*}{qwen2vl} & BLEU & 0.0597 & 0.0679 & 0.0526 & 0.0570 \\
 & ROUGE-L & 0.3300 & 0.3238 & 0.3174 & 0.2990 \\
 & BERTScore & 0.3107 & 0.2980 & 0.3052 & 0.2932 \\
\midrule
\multirow{3}{*}{llavanext} & BLEU & 0.0052 & 0.0205 & 0.0113 & 0.0239 \\
 & ROUGE-L & 0.1663 & 0.2038 & 0.1718 & 0.2257 \\
 & BERTScore & 0.0700 & 0.1066 & 0.0727 & 0.1540 \\
\midrule
\multirow{3}{*}{llavaov} & BLEU & 0.0226 & 0.0282 & 0.0215 & 0.0236 \\
 & ROUGE-L & 0.1390 & 0.1459 & 0.1329 & 0.1286 \\
 & BERTScore & 0.0066 & 0.0094 & -0.0400 & -0.0327 \\
\midrule
\multirow{3}{*}{videollama} & BLEU & 0.0504 & 0.0624 & 0.0339 & 0.0411 \\
 & ROUGE-L & 0.2643 & 0.2870 & 0.2326 & 0.2537 \\
 & BERTScore & 0.2573 & 0.2552 & 0.2049 & 0.2391 \\
\midrule
\multirow{3}{*}{llama3ft} & BLEU & \textbf{0.1073} & \textbf{0.1306} & \textbf{0.0987} & \textbf{0.1068} \\
 & ROUGE-L & \textbf{0.4045} & \textbf{0.4279} & \textbf{0.3845} & \textbf{0.3927} \\
 & BERTScore & \textbf{0.4622} & \textbf{0.4669} & \textbf{0.4279} & \textbf{0.4319} \\
\bottomrule
\end{tabularx}
\end{table}

We demonstrate a systematic framework for characterising the depth of understanding of AI systems in the cooking domain. Though today's AI systems are very promising for many tasks, there is a good amount of work leftout in developing skills required for understanding fine and specialized skills, as in domains like cooking.

%% file: content/applications_and_conclusion.tex
\begin{figure*}[htbp] 
    \centering
    \includegraphics[width= 1.8 \columnwidth]{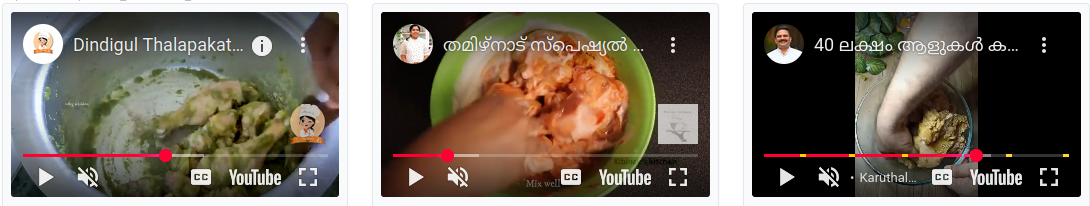} 
    \caption{Example of skill-based video retrieval for the query \textit{``marinating chicken''}. 
The system returns short, timestamped clips from multiple biryani videos where the marination step is visually identified, enabling direct access to semantically relevant moments rather than full unindexed videos.}
\Description{
    Three thumbnail images of YouTube videos are shown. 
    The first video (left) depicts chicken pieces being mixed with a greenish marinade in a pot (Dindigul Thalapakatti style). 
    The second video (centre) shows raw chicken coated in a reddish marinade inside a green bowl. 
    The third video (right) depicts chicken being marinated by hand in a dark brown paste inside a glass bowl. 
    These highlight visual diversity in marination steps across recipes.}
    \label{fig:action_retreival}
\end{figure*}

\section{Discussions}

\paragraph{\bf Application in Skill-Based Video Retrieval}

Beyond full-recipe visualisation, our dataset supports targeted instructional search within and across videos. For instance, if a user is interested in understanding how to marinate chicken—a critical step in many \biryani variants—they can retrieve all video segments across the dataset that involve marination actions. These segments are sourced from different videos but are uniformly timestamped and labelled using our alignment framework. Figure~\ref{fig:action_retreival} presents an example frame retrieved from a marination segment. Unlike traditional video search engines, which return entire videos without pinpointing where the relevant action occurs, our approach enables direct navigation to semantically aligned moments within the video corpus.

Our work opens up many more potential applications in cooking:
\begin{itemize}
    \item Understanding and documenting the rich cultural heritage of the country, enabling its transfer and preservation.
    \item We hope the deeper video understanding presented here could lead to educational tool and cooking assistants, who can provide contextual assistance with speech and language when integrated with an ego-centric vision.
\end{itemize}


\subsection{Summary}
In this work, we presented a systematic computational study of \biryani preparation videos from across India. We aimed to understand how fine-grained procedural differences manifest in culturally rich cooking practices. We curated the first large-scale \Biryani Cooking Video Dataset, comprising 120 high-quality YouTube videos spanning 12 distinct regional styles. Building on recent advances in vision–language models (VLMs), we developed a multi-stage framework for temporal segmentation and multimodal alignment between visual content, narration, and canonical recipe text.

We used this aligned representation to introduce a video comparison pipeline that identifies and explains procedural differences between regional variants, enabling interpretable cross-recipe analysis. We further constructed a multi-tier question–answer benchmark to evaluate VLMs on procedural video understanding tasks ranging from localised recognition to multi-video reasoning. Our experiments benchmarked several state-of-the-art VLMs under both zero-shot and fine-tuned settings, highlighting the potential of domain adaptation for structured multimodal reasoning.

Beyond its immediate results, this work provides a foundation for a new class of video understanding benchmarks that combine cultural specificity with fine-grained procedural analysis. The dataset, prompts, and annotations will be released to facilitate reproducibility and further research. Future directions include expanding the scope to other culturally significant dishes, improving alignment robustness in the presence of noisy narration, and developing more efficient VLM prompting strategies for long-form video. 

\paragraph{\bf Acknowledgements}

We acknowledge and appreciate the support of Google Research / AI in this project.